\documentclass[journal,onecolumn]{IEEEtran}

\usepackage{epsfig}
\usepackage{graphicx}
\usepackage{amsmath}
\usepackage{amssymb}
\usepackage{commath}
\usepackage{bbding}
\usepackage{cite}
\usepackage{cleveref}
\usepackage{booktabs,multirow}
\usepackage{times}
\usepackage{epstopdf}

\usepackage{mathtools}

\begin{document}

\title{\huge Phase-Shifting Separable Haar Wavelets and Applications}
%
%

\author{Mais~Alnasser\thanks{Mais Alnasser was with the Department of Computer Science, University of Central Florida, Orlando, FL, 32816 USA at the time this project was conducted. (e-mail: nasserm@cs.ucf.edu).} and Hassan~Foroosh\thanks{Hassan Foroosh is with the Department of Computer Science, University of Central Florida, Orlando, FL, 32816 USA (e-mail: foroosh@cs.ucf.edu).}
}

\maketitle

\begin{abstract}
This paper presents a new approach for tackling the shift-invariance
problem in the discrete Haar domain, without trading off any of its
desirable properties, such as compression, separability,
orthogonality, and symmetry. The paper presents several key
theoretical contributions. First, we derive closed form expressions
for phase shifting in the Haar domain both in partially decimated and
fully decimated transforms. Second, it is shown that the wavelet
coefficients of the shifted signal can be computed solely by using the
coefficients of the original transformed signal. Third, we derive
closed-form expressions for non-integer shifts, which have not been
previously reported in the literature. Fourth, we establish the
complexity of the proposed phase shifting approach using the derived
analytic expressions. As an application example of these results, we
apply the new formulae to image rotation and interpolation, and
evaluate its performance against standard methods.
\end{abstract}


\section{Introduction}

The wavelet transform has been playing an ever
increasing important role in the modeling and analysis of a wide range
of problems in science and engineering. In signal and image
processing, wavelets have been particularly instrumental in methods of
constructing ``optimal'' basis that are often used in various image
processing and computer vision applications, such as shape/scene description and classification  \cite{liu2015sparse,ChangK93TexAnalysis,saito94LDB,saito95LOB,DoVetterli-rotation,Do99invariant,do00texture,Cakmakci_etal_2008,Cakmakci_etal_2008_2,Zhang_etal_2015,Lotfian_Foroosh_2017,Morley_Foroosh2017,Ali-Foroosh2016,Ali-Foroosh2015,Einsele_Foroosh_2015,ali2016character,Cakmakci_etal2008,damkjer2014mesh,Junejo_etal_2013,bhutta2011selective,junejo1dynamic,ashraf2007near,Junejo_etal_2007,Junejo_Foroosh_2008,Sun_etal_2012,junejo2007trajectory,sun2011motion,Ashraf_etal2012,sun2014feature,Junejo_Foroosh2007-1,Junejo_Foroosh2007-2,Junejo_Foroosh2007-3,Junejo_Foroosh2006-1,Junejo_Foroosh2006-2,ashraf2012motion,ashraf2015motion,sun2014should},  scene content modeling \cite{Junejo_etal_2010,Junejo_Foroosh_2010,Junejo_Foroosh_solar2008,Junejo_Foroosh_GPS2008,junejo2006calibrating,junejo2008gps,Tariq_etal_2017,Tariq_etal_2017_2,tariq2013exploiting,tariq2015feature,tariq2014scene}, image restoration and denoising \cite{coifman94adapted,coifman95translationinvariant,donoho95denoising,Figueiredo-etal07b,KrimMallatBestBasis,Hu_etal_IBR2012,Foroosh_2000,Foroosh_Chellappa_1999,Foroosh_etal_1996,Cao_etal_2015,berthod1994reconstruction,shekarforoush19953d,lorette1997super,shekarforoush1998multi,shekarforoush1996super,shekarforoush1995sub,shekarforoush1999conditioning,shekarforoush1998adaptive,berthod1994refining,shekarforoush1998denoising,bhutta2006blind,jain2008super,shekarforoush2000noise,shekarforoush1999super,shekarforoush1998blind}, video content modeling \cite{Shen_Foroosh_2009,Ashraf_etal_2014,Ashraf_etal_2013,Sun_etal_2015,shen2008view,sun2011action,ashraf2014view,shen2008action,shen2008view-2,ashraf2013view,ashraf2010view,boyraz122014action,Shen_Foroosh_FR2008,Shen_Foroosh_pose2008,ashraf2012human},  image alignment \cite{Foroosh_etal_2002,Foroosh_2005,Balci_Foroosh_2006,Balci_Foroosh_2006_2,Alnasser_Foroosh_2008,Atalay_Foroosh_2017,Atalay_Foroosh_2017-2,shekarforoush1996subpixel,foroosh2004sub,shekarforoush1995subpixel,balci2005inferring,balci2005estimating,foroosh2003motion,Balci_Foroosh_phase2005,Foroosh_Balci_2004,foroosh2001closed,shekarforoush2000multifractal,balci2006subpixel,balci2006alignment,foroosh2004adaptive,foroosh2003adaptive}, tracking and object pose estimation \cite{Shu_etal_2016,Milikan_etal_2017,Millikan_etal2015,shekarforoush2000multi,millikan2015initialized},  camera motion quantification and calibration \cite{Cao_Foroosh_2007,Cao_Foroosh_2006,Cao_etal_2006,Junejo_etal_2011,cao2004camera,cao2004simple,caometrology,junejo2006dissecting,junejo2007robust,cao2006self,foroosh2005self,junejo2006robust,Junejo_Foroosh_calib2008,Junejo_Foroosh_PTZ2008,Junejo_Foroosh_SolCalib2008,Ashraf_Foroosh_2008,Junejo_Foroosh_Givens2008,Lu_Foroosh2006,Balci_Foroosh_metro2005,Cao_Foroosh_calib2004,Cao_Foroosh_calib2004,cao2006camera}, and image-based rendering (IBR) \cite{Cao_etal_2005,Cao_etal_2009,shen2006video,balci2006real,xiao20063d,moore2008learning,alnasser2006image,Alnasser_Foroosh_rend2006,fu2004expression,balci2006image,xiao2006new,cao2006synthesizing}, to name a few. However, a major drawback restricting the use of such
methods is the lack of shift-invariance. For example, in the case of
de-noising, Gibbs phenomenon in the neighborhood of discontinuities is
attributed to the lack of shift-invariance of the wavelet basis
\cite{coifman95translationinvariant}. An image transform is
shift-invariant if the total energy of the coefficients in any subband
is invariant to translations of the original image. It can be thus
readily verified that the fastest and the most compact formulations -
i.e. the classical fully decimated real wavelet transforms - suffer
from the lack of shift-invariance. Additional properties that are
often desired in many applications of wavelets include separability,
orthogonality and symmetry.

There has been two trends in responding to the shift-invariance
requirement. The earlier literature has been focusing on modifying the
classical real wavelets to enforce shift-invariance, while attempting
to preserve other desired properties. This approach was rediscovered
by various authors independently, and bears different names such as
{\em algorithme {\`a} trous}
\cite{HolschneiderSigAnaWT,CombesImpAlgAtrous,Mallat98wavelet},
redundant wavelets \cite{Burrus97IntroWT} and undecimated wavelets
\cite{lang96noise} to name a few. The major drawbacks of this
approach, of course, are the undesirable side-effect of overly
redundant representation and the high computational cost, since each
set of coefficients contains the same number of samples as the input
signal. This level of redundancy essentially defeats the purpose of
designing wavelets for compression and coding, which take advantage of
the localization properties of wavelets as opposed to the
shift-invariant Fourier basis.

In order to alleviate these side-effects, more recently a second
approach has been investigated in the literature that attempts to
directly construct shift-invariant wavelets. This line of research has
led to a new class of wavelets with complex coefficients. Few examples
are the Gabor wavelets for texture processing \cite{ManjunathGabor},
harmonic wavelets for vibration and acoustic analysis
\cite{NewlandHW,NewlandHWva} and the Complex Wavelet Transform (CWT)
for motion estimation \cite{MagareyMotionEstimation}. In addition to
shift-invariance, one particular advantage of complex wavelets is
directionality that is similar to the steerable pyramids
\cite{simoncelli95steerable}. Complex wavelets prove to be useful in
solving the shift-invariance problem without compromising many other
properties. However, their major drawbacks are lack of speed and often
also poor inversion properties. A more successful attempt in this
category is perhaps the dual-tree complex wavelet transform (DT-CWT)
and its variations \cite{RivazInvariantComplex,RivazK00}.  Although,
DT-CWT provides a good trade-off between fully decimated wavelets and
the redundant wavelet transform, it does so by trading off the
compression capabilities and computational time of the classical real
wavelets.

In this paper, we initiate and investigate a third line of approach to
tackling the shift-invariance problem. Instead of modifying a
classical wavelet or introducing a new complex wavelet, our goal is to
determine in what way the wavelet coefficients in a fully decimated
transform are related to those of a shifted signal. Of course such
relation would be wavelet-dependent and may not be a straightforward
relation as in redundant wavelets, where the shift in the input
results in a shift in the output. The key idea is that as long as the
relation is known, one can tackle shift-invariance, since all the
coefficients of a shifted signal can be mapped to those of the
original signal. On the other hand, shift-invariance is tackled
without compromising speed and compression properties.  Furthermore,
establishing the explicit and direct relations between the
coefficients of a signal and its shifted version, would allow us to
perform compressed domain processing of signals or images without
requiring a chain of forward and backward transforms. This is
particularly of interest in applications such as data compression and
progressive transmission, or more recent applications in compressed
sensing \cite{Castro-etal06,Figueiredo-etal07a,Haupt-Nowak05}. Our
focus in this paper is on the standard Haar wavelet transform due to
additional desirable properties of separability and symmetry.

We present a solution to phase-shift the Haar coefficients in the
transform domain solely using the available coefficients of the
unshifted transformed signal, which we refer to as the 0-shift signal.
Our solution generalizes readily to an N-dimensional signal due to
separability. We also show how our solution can be extended to
non-integer phase shifts. To demonstrate the power of the proposed
approach and to evaluate it, we performed extensive experiments on the
problem of accurate image rotation
\cite{DBLP:journals/tip/UnserTY95}. The remaining of this paper is
organized as follows: In the next section, we introduce the notations
and briefly describe the Haar transform tree. The following two
sections will then derive our expressions for describing the explicit
relations between the Haar coefficients of a 0-shift and shifted
signal for both fully and partially transformed signals. These results
are then extended for sub-pixel shifting, followed by full evaluation
and testing of the results on image rotation and interpolation
problems. The paper concludes with a brief discussion and some remarks
on the proposed new ideas.


\section{The Haar Transform Tree}

\begin{figure*}[!t]
\centering
\includegraphics[width=5in]{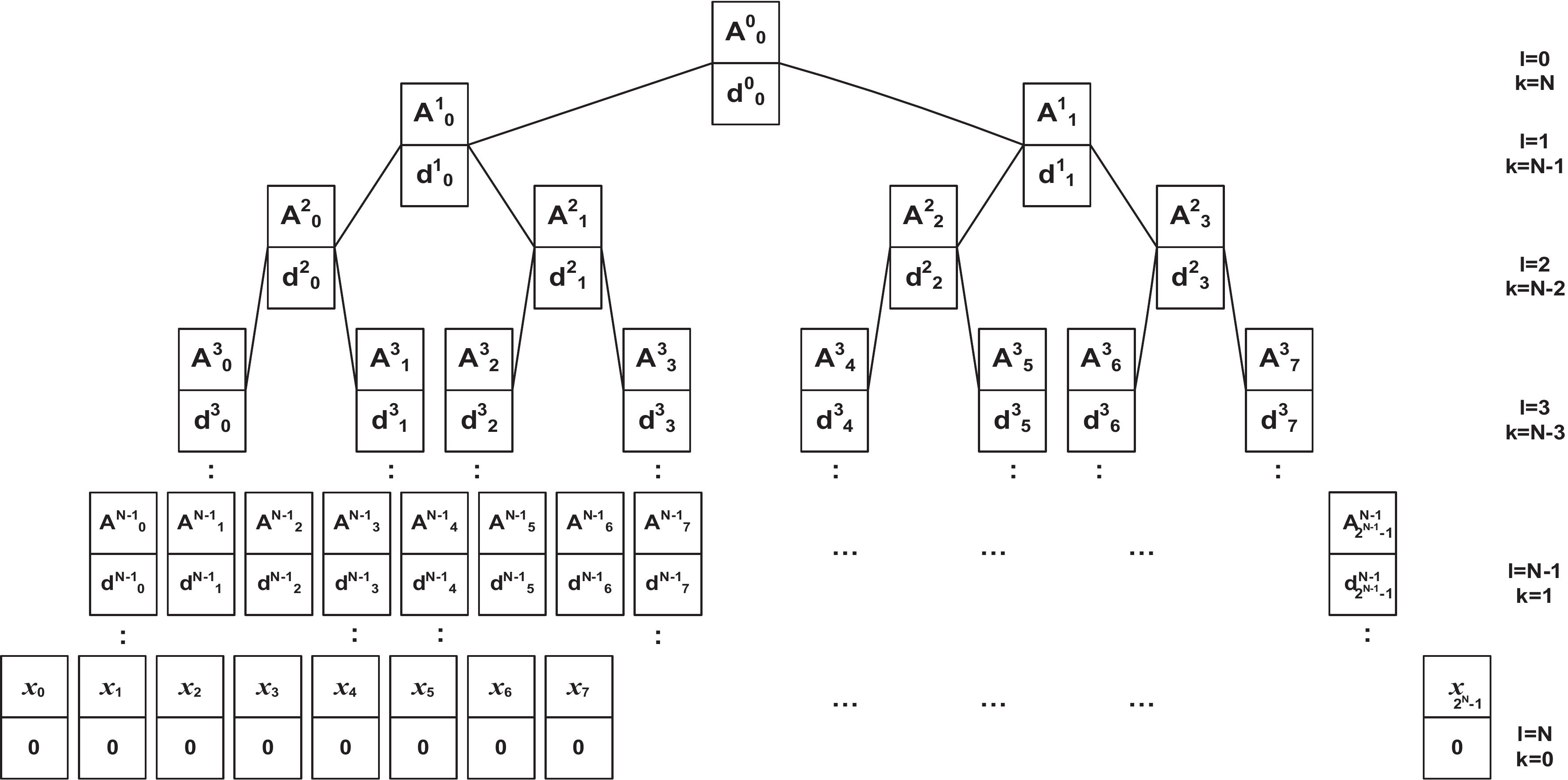}
\caption{The Haar coefficients tree contains at each level the blur
and the detail coefficients $A^l_i$ and $d^l_i$, respectively. The
Haar transform of the one-dimensional signal $x(n)$ is composed of the
dc value $A^0_0$ and the detail coefficients $d^l_i$, where
$l=0,...,2^{N-1}$ and $i=0,...,2^l-1$. The blur coefficients $A^l_i$
at each level are used to derive the analytic expressions for
phase-shifting the signal $x(n)$, but the final expressions are
independent of the blur coefficients. Note that the leaves of the
tree, which are composed of the signal $x(n)$ can be considered as the
blur coefficients at level $N$, namely $A^N_i$.} \label{fig:oneDtree}
\end{figure*}

Let $x(n)$ be a one-dimensional signal of size $2^N$, where $N$ is a
positive integer. The Haar transform of $x(n)$, namely $H(x(n))$,
has the form:
\begin{equation} H(x(n)) = \{A_0^0, d_0^0, d^1_0, d^1_1, ..., d^l_i, ..., d^{N-1}_0, ..., d^{N-1}_{2^{N-1}-1}\}
\end{equation}

such that $A_0^0$ is the dc value of the signal and $d^l_i$ is the
$i^{t\!h}$ detail coefficient at level $l$, where $l=0,...,N-1$ and
$i=0,...,2^l-1$.

Transforming a signal using Haar wavelets can be easily done by
successively convolving the blurred part of the signal by box and
differencing filters until the signal is fully transformed (see for
instance \cite{VetterliJelena} for more details).

We choose to express the Haar transformation using a tree as in Fig.
\ref{fig:oneDtree}. The tree is constructed of $N$ levels with $x(n)$
residing at the leaves, i.e. the $N^{\mbox{th}}$ level. The
$i^{\mbox{th}}$ node at level $l$ in the tree can be made to hold the
0-shift $i^{\mbox{th}}$ blur and detail coefficients, $A^l_i$ and
$d^l_i$, respectively, where $l=0,...,N-1$ and $i=0,...,2^l-1$.

Each level in the tree corresponds to a reduction step $k=1,...,N$,
with the untransformed original signal corresponding to $k=0$. The
signal is partially transformed with $k$ reduction steps if $0<k<N$
and is said to be fully transformed if $k=N$. At each reduction level
$k$, one obtains the partially transformed signal
$H^k(x(n))$. $H^k(x(n))$ is composed of the blur coefficients at level
$k$ followed by the detail coefficients at the same level and all
subsequent reduction levels that are less than $k$ and greater than
1. That is:
\begin{eqnarray}
H^k(x(n)) &=& \{A_0^{N-k}, ..., A_{2^{N-k}-1}^{N-k}, d_0^{N-k}, ...,
d_{2^{N-k}-1}^{N-k},..., \nonumber \\
&& d^l_i, ..., d^{N-1}_0, ..., d^{N-1}_{2^{N-1}-1}\}
\end{eqnarray} Where, $l=N-k,...,N-1$ and $i=0,...,2^l$. Note that
$H^N(x(n))=H(x(n))$ is the fully transformed signal.

We use the tree to examine the behavior of the detail coefficients
with respect to shifting. Note that we can denote $x(i)$ as $A^N_i$,
in which case $d^N_i=0$. By using this notation, $l$ now has the range
$0,...,N$. Also, note that the blur coefficient $A^{l}_{i}$ is related
to its parent at level $l-1$ by the following
relation:
\begin{eqnarray} 
A^{l}_{i} &=& A^{l-1}_{i/2}\;+\;d^{l-1}_{i/2}\;,\mbox{$i$ is even}
\nonumber\\ &&
A^{l-1}_{\lfloor{i/2}\rfloor}-d^{l-1}_{\lfloor{i/2}\rfloor},\mbox{$i$
is odd} \label{eq:AArel}
\end{eqnarray}

Now, let $D^l_i$ be the difference between the dc value at the root of
the tree $A^0_0$ and the blur coefficient $A^l_i$. Then
\begin{equation} 
A^l_i=A^0_0+D^l_i \label{eq:ADrel}
\end{equation}

By substituting (\ref{eq:ADrel}) in (\ref{eq:AArel}), $D^l_i$ can be
computed recursively solely in terms of the detail coefficients
using the following relation:
\begin{eqnarray}
D^l_i & = & D^{l-1}_{i/2}\;\;+\;d^{l-1}_{i/2}\;,\mbox{$i$ is even} \nonumber\\
& & {}
D^{l-1}_{\lfloor{i/2}\rfloor}-d^{l-1}_{\lfloor{i/2}\rfloor}, \mbox{$i$ is odd} \nonumber \\
D^0_0 & = & 0 \label{eq:DDrel}
\end{eqnarray}

It can be verified that $D^l_i$ can be computed recursively with a
complexity of $O(l)$ for fully-transformed signals, which in itself is
very cheap, or be simply tabulated for even a faster retrieval. Also,
note that for partially transformed signals, a combination of
(\ref{eq:DDrel}) and (\ref{eq:ADrel}) has to be used to evaluate
$D^l_i$:
\begin{eqnarray}
l=N-k: & & \nonumber\\
D^l_i & = & A^l_i-A^0_0 \nonumber\\
l>N-k: & & \nonumber\\
D^l_i & = & D^{l-1}_{i/2}\;\;+\;d^{l-1}_{i/2}\;,\mbox{$i$ is even} \nonumber\\
& & D^{l-1}_{\lfloor{i/2}\rfloor}-d^{l-1}_{\lfloor{i/2}\rfloor},
\mbox{$i$ is odd} \label{eq:DDrel_part}
\end{eqnarray}
The complexity for the above equation is even less than that of
(\ref{eq:DDrel}) because the recursion needs to go a maximum depth
of $k$ rather than a maximum depth of $N$. In other words, the
complexity for the above equation is $O(k-l)$

At level $N-k$, there are $2^k$ non-redundant coefficient sets each
of size $2^{N-k}$ \cite{sarisarraf97shiftinvariant}, where $k=1,...,N$. A shift $s=0,...,2^N-1$ can be one of the following possibilities:
\begin{itemize}
\item A shift that is divisible by $2^k$. 
\item An odd shift. 
\item An even shift that is not divisible by $2^k$.
\end{itemize}

In the following sections, we first analyze the behavior of the detail
coefficients based on the above three possibilities for a fully
transformed signal. We then analyze the behavior of the blur
coefficients for signals that are partially transformed. The final
analytic solutions that we provide are capable of evaluating the
coefficients of the shifted signal solely using the original
coefficients of the 0-shift signal, which is the goal of our paper.

\section{Shifting Fully Transformed Signals}
\subsection{Shifting by a Multiple of $2^k$} 
This is the simplest case. A shift $s$ in the discrete domain that is
equal to $2^ku$ is a circular shift of the 0-shift detail coefficients
at level $N-k$ by $u$, that is,
\begin{equation}
d^{N-k}_{i_{new}} = d^{N-k}_{(i+u)\%2^{N-k}} , \;\;\;\; k=1,...,N
\end{equation}
where $0\leq u \leq 2^{N-k}-1$ and \% is the mod operation. Note that
for levels $N-(k-1), N-(k-2),..., N-1$ a shift of $2^ku$ of the
original signal is a circular shift of the coefficients at those
levels by $2u, 2^2u,..., 2^{k-1}u$, respectively. In other words, a
shift of $2^ku$ of the original signal shifts the coefficients at
level $N-k$ by $u$, while shifting the coefficients at level $N-(k-1)$
by twice as much, and the coefficients at level $N-(k-2)$ by four
times as much and so on.

\subsection{Shifting by an Odd Amount}
By examining the tree in figure (\ref{fig:oneDtree}), we notice
that:
\begin{eqnarray} d^{N-k}_{i_{new}} \!\!\!& = &\!\!\!
(( x_{(2^{k}i+s)\%2^N} + ... + x_{(2^{k-1}(2i+1)+s-1)\%2^N} ) \nonumber\\
\!\!\!& - &\!\!\! ( x_{(2^{k-1}(2i+1)+s)\%2^N} + ... +
x_{(2^k(i+1)+s-1)\%2^N} ))/2^k\nonumber\\ \label{eq:xn1}
\end{eqnarray} 
In other words, $d^{N-k}_{i_{new}}$ is the sum of the leaves shifted
into its left branch minus the leaves shifted into its right branch
divided by $2^k$. To simplify the above equation, we set the indices
as follows:
\begin{eqnarray*}
i_1&=&2^{k}i+s\\
i_2&=&2^{k-1}(2i+1)+s\\
i_3&=&2^{k}(i+1)+s
\end{eqnarray*}
Using the notation $A^N_i$ for $x_i$, (\ref{eq:xn1}) now becomes:
\begin{eqnarray}
d^{N-k}_{i_{new}} & = & (( A^N_{i_1\%2^N} + ... + A^N_{(i_2-1)\%2^N} ) \nonumber\\
& - & ( A^N_{i_2\%2^N} + ... + A^N_{(i_3-1)\%2^N} ))/2^k
\label{eq:Nlevel}
\end{eqnarray}

Substituting (\ref{eq:ADrel}) and then (\ref{eq:DDrel}) in
(\ref{eq:Nlevel}) and canceling out the $A^0_0$'s, the relation for
computing $d^{N-k}_{i_{new}}$ for a shift $s$ that is
odd becomes:
\begin{eqnarray}
\!\!\!\!\!\!\!\!\!d^{N-k}_{i_{new}} \!\!\!\!\!& = &\!\!\!\!\!
(D^{N-1}_{i_1\%2^{N-1}} +
2\sum_{m=i_1+1}^{i_2-1}{D^{N-1}_{m\%2^{N-1}}}
- 2\sum_{m=i_2+1}^{i_3-1}{D^{N-1}_{m\%2^{N-1}}} \nonumber\\
\!\!\!\!\!\!\!\!& - &\!\!\!\!\! D^{N-1}_{i_3\%2^{N-1}}\!-\!
d^{N-1}_{i_1\%2^{N-1}}\!\! +\!\!
2d^{N-1}_{i_2\%2^{N-1}}\! - \!d^{N-1}_{i_3\%2^{N-1}})/2^k\nonumber\\
\mbox{where,}\nonumber\\
i_1&=&2^{k-1}i+\lfloor{s/2}\rfloor\nonumber\\
i_2&=&2^{k-2}(2i+1)+\lfloor{s/2}\rfloor\nonumber\\
i_3&=&2^{k-1}(i+1)+\lfloor{s/2}\rfloor
\end{eqnarray}
Note that for $k=1$, $i_2$ would be a non-integer value, in which case
we must set $d^{N-1}_{i_2\%2^{N-1}}$ to 0.

\subsection{Shifting by an Even Amount that is Not Divisible by $2^k$}

In this case, $s$ is divisible by $2^t$, for $1 \leq t \leq k-1$ and
$2^t$ is the highest power of 2 by which $s$ is divisible.  This
allows us to let $s=2^tu$, where $0\leq u \leq 2^{N-t}-1$. This means
that the coefficients at levels $N-1,...,N-t$ follow the first
case. In other words, the 0-shift coefficients at levels
$N-1,N-2,...,N-t$ are circularly shifted by $2^{t-1}u,
2^{t-2}u,...,u$, respectively. Since $2^t$ is the highest power of 2
by which $s$ is divisible, $u$ must be odd. This allows us to treat
this case as an odd shift of the blur details at level $N-t$. In other
words, at level $N-k$, $d^{N-k}_{i_{new}}$ can be evaluated using the
following modification of equation (\ref{eq:Nlevel}):
\begin{eqnarray}
d^{N-k}_{i_{new}} & = & (( A^{N-t}_{i_1\%2^{N-t}} + ... + A^{N-t}_{(i_2)-1\%2^{N-t}} ) \nonumber\\
& - & ( A^{N-t}_{i_2\%2^{N-t}} + ... + A^{N-t}_{(i_3-1)\%2^{N-t}}
))/2^{k-t}\nonumber\\
\mbox{where,}\nonumber\\
i_1&=&2^{k-t}i+s/2^t\nonumber\\
i_2&=&2^{k-t-1}(2i+1)+s/2^t\nonumber\\
i_3&=&2^{k-t}(i+1)+s/2^t\label{eq:NtLevel}
\end{eqnarray}
Following the same steps, the above can be rewritten as:
\begin{eqnarray}
d^{N-k}_{i_{new}} \!\!\!\!\!& = &\!\!\!\!\!
(D^{N-t-1}_{i_1\%2^{N-t-1}} +
2\sum_{m=i_1+1}^{i_2-1}{D^{N-t-1}_{m\%2^{N-t-1}}} \nonumber\\
& - &\!\!\!\! 2\sum_{m=i_2+1}^{i_3-1}{D^{N-t-1}_{m\%2^{N-t-1}}} -
D^{N-t-1}_{i_3\%2^{N-t-1}} \nonumber\\
& - & \!\!\!\! d^{N-t-1}_{i_1\%2^{N-t-1}} + 2d^{N-t-1}_{i_2\%2^{N-t-1}} - d^{N-t-1}_{i_3\%2^{N-t-1}})/2^{k-t}\nonumber\\
\mbox{where,}\nonumber\\
i_1&=&2^{k-t-1}i+\lfloor{s/2^{t+1}}\rfloor\nonumber\\
i_2&=&2^{k-t-2}(2i+1)+\lfloor{s/2^{t+1}}\rfloor\nonumber\\
i_3&=&2^{k-t-1}(i+1)+\lfloor{s/2^{t+1}}\rfloor\nonumber\\
\mbox{and,   }\;\;\;\;&&\nonumber\\
d^{N-1}_{i_2\%2^{N-1}}&=&0 \mbox{, if $i_2$ is non-integer.}
\end{eqnarray}

Note that the second case is the same as the third case when $t=0$.
That leaves us with the following formula:
\begin{eqnarray}
k>t:\nonumber\\
d^{N-k}_{i_{new}} \!\!\!\!\!& = &\!\!\!\!\!
(D^{N-t-1}_{i_1\%2^{N-t-1}} +
2\sum_{m=i_1+1}^{i_2-1}{D^{N-t-1}_{m\%2^{N-t-1}}} \nonumber\\
& - &\!\!\!\! 2\sum_{m=i_2+1}^{i_3-1}{D^{N-t-1}_{m\%2^{N-t-1}}} -
D^{N-t-1}_{i_3\%2^{N-t-1}} \nonumber\\
& - & \!\!\!\! d^{N-t-1}_{i_1\%2^{N-t-1}} + 2d^{N-t-1}_{i_2\%2^{N-t-1}} - d^{N-t-1}_{i_3\%2^{N-t-1}})/2^{k-t}\nonumber\\
k \leq t: \nonumber\\
d^{N-k}_{i_{new}} & = & d^{N-k}_{(i+s/2^k)\%2^{N-k}}\nonumber\\
\mbox{where,}\;\;\;\nonumber\\
i_1&=&2^{k-t-1}i+\lfloor{s/2^{t+1}}\rfloor\nonumber\\
i_2&=&2^{k-t-2}(2i+1)+\lfloor{s/2^{t+1}}\rfloor\nonumber\\
i_3&=&2^{k-t-1}(i+1)+\lfloor{s/2^{t+1}}\rfloor\nonumber\\
\mbox{and,   }\;\;\;\;&&\nonumber\\
d^{N-t-1}_{i_2\%2^{N-1}}&=&0 \mbox{, if $i_2$ is a
non-integer}\label{eq:dfinal}
\end{eqnarray}

The above relation can now be used to evaluate the new detail
coefficients of the Haar transform at all different levels after any
shift $s=0,...,2^N-1$ using only the coefficients of the 0-shift
signal. The worst case complexity for evaluating $d^{N-k}_{i_{new}}$
using (\ref{eq:dfinal}) is \emph{O}$(\log(L))$, where $L$ is the size
of the signal $x(n)$ (see the complexity analysis section for more
details).

\section{Shifting Partially Transformed Signals}
Depending on the application, the original signal might not be fully
transformed. As we mentioned earlier, a signal that has $k$
degrees of reduction has the form:
\begin{eqnarray*}
H^k(x(n)) &=& \{A_0^{N-k}, ..., A_{2^{N-k}-1}^{N-k}, d_0^{N-k}, ...,
d_{2^{N-k}-1}^{N-k},..., \\
&& d^l_i, ..., d^{N-1}_0, ..., d^{N-1}_{2^{N-1}-1}\}
\end{eqnarray*} 
Where, $1\leq k\leq N-1$, $l=N-k,...,N-1$ and $i=0,...,2^l$.

A signal that is partially transformed is composed of both blur
coefficients and detail coefficients. Equation (\ref{eq:dfinal})
shows how to evaluate the detail coefficients of a fully transformed
shifted signal, which also applies to evaluating the detail
coefficients of a partially transformed signal. In this section we
show how to evaluate the blur coefficients at reduction step $k$ for
a signal that has been decomposed $k$ times and shifted by the
integer amount $s$ in the time domain.

\subsection{Shifting by a Multiple of $2^k$}
Similar to evaluating the detail coefficients case, a shift $s$ in
the discrete domain that is equal to $2^ku$ is a circular shift
of the 0-shift blur coefficients at level $N-k$ by $u$, that is,
\begin{equation}
A^{N-k}_{i_{new}} = A^{N-k}_{(i+u)\%2^{N-k}} , \;\;\;\; k=1,...,N-1
\end{equation}
where $0\leq u \leq 2^{N-k}-1$.

\subsection{Shifting by an Odd Amount}
By examining the tree in figure (\ref{fig:oneDtree}), we notice
that:
\begin{eqnarray} A^{N-k}_{i_{new}} \!\!\!& = &\!\!\!
(( x_{(2^{k}i+s)\%2^N} + ... + x_{(2^{k-1}(2i+1)+s-1)\%2^N} ) \nonumber\\
\!\!\!& + &\!\!\! ( x_{(2^{k-1}(2i+1)+s)\%2^N} + ... +
x_{(2^k(i+1)+s-1)\%2^N} ))/2^k\nonumber\\ \label{eq:xn}
\end{eqnarray} 
In other words, $A^{N-k}_{i_{new}}$ is the sum of the leaves shifted
into its left branch plus the leaves shifted into its right branch
divided by $2^k$. To simplify the above equation, we use only the
starting and ending coefficients and we also use the notation $A^N_i$
for $x_i$:
\begin{eqnarray}
A^{N-k}_{i_{new}} & = & ( A^N_{i_1\%2^N} + ... + A^N_{(i_2-1)\%2^N}
)/2^k \label{eq:Nlevel_part}
\end{eqnarray}

Where,
\begin{eqnarray*}
i_1&=&2^{k}i+s\\
i_2&=&2^{k}(i+1)+s
\end{eqnarray*}

Substituting (\ref{eq:AArel}) in the above, we get
\begin{eqnarray}
A^{N-k}_{i_{new}} & = & ( A^0_0 + D^N_{i_1\%2^N} + ... + A^0_0 +
D^N_{(i_2-1)\%2^N}
)/2^k 
\end{eqnarray}
The number of $A^0_0$'s is equal to the number of coefficients
$A^l_i$ being summed, which is equal to $2^k$. We factor out
$A^0_0$:
\begin{eqnarray}
A^{N-k}_{i_{new}} & = & A^0_0 + (D^N_{i_1\%2^N} + ... +
D^N_{(i_2-1)\%2^N}
)/2^k 
\end{eqnarray}

Substituting (\ref{eq:DDrel}) and simplifying, we get the analytic
solution for evaluating $A^{N-k}_{i_{new}}$ under an odd shift $s$:
\begin{eqnarray}
\!\!\!\!\!\!\!\!\!A^{N-k}_{i_{new}} \!\!\!\!\!& = &\!\!\!\!\! A^0_0
+ (D^{N-1}_{i_1\%2^{N-1}} +
\sum_{m=i_1+1}^{i_2-1}D^{N-1}_{m\%2^{N-1}} +
D^{N-1}_{i_2\%2^{N-1}}\nonumber\\
&& - d^{N-1}_{i_1\%2^{N-1}} + d^{N-1}_{i_2\%2^{N-1}})/2^k
\nonumber\\
\mbox{where,}\nonumber\\
i_1&=&2^{k-1}i+\lfloor{s/2}\rfloor\nonumber\\
i_2&=&2^{k-1}(i+1)+\lfloor{s/2}\rfloor
\end{eqnarray}

\subsection{Shifting by an Even Amount that is Not Divisible by $2^k$} 

For a shift $s=2^tu$, where $0\leq u \leq 2^{N-t}-1$ and $t<k$, we can
treat this case as an odd shift of the coefficients at level $N-t$,
which is similar to what we did in evaluating the detail coefficients
under a shift $s=2^tu$. $A^{N-k}_{i_{new}}$ can now be evaluated using
the following equation:
\begin{eqnarray}
A^{N-k}_{i_{new}} & = & ( A^{N-t}_{i_1\%2^{N-t}} + ... +
A^{N-t}_{(i_2-1)\%2^{N-t}} )/2^{k-t} 
\end{eqnarray}

Proceeding as we did in the odd shift case, we get the following
solution:

\begin{eqnarray}
\!\!\!\!\!\!\!\!\!A^{N-k}_{i_{new}} \!\!\!\!\!& = &\!\!\!\!\! A^0_0
+ (D^{N-t-1}_{i_1\%2^{N-t-1}} +
\sum_{m=i_1+1}^{i_2-1}D^{N-t-1}_{m\%2^{N-t-1}} +
D^{N-t-1}_{i_2\%2^{N-t-1}}\nonumber\\
&& - d^{N-t-1}_{i_1\%2^{N-t-1}} +
d^{N-t-1}_{i_2\%2^{N-t-1}})/2^{k-t}
\nonumber\\
\mbox{where,}\nonumber\\
i_1&=&2^{k-t-1}i+\lfloor{s/2^{t+1}}\rfloor\nonumber\\
i_2&=&2^{k-t-1}(i+1)+\lfloor{s/2^{t+1}}\rfloor
\end{eqnarray}

Combining the three cases, the final result becomes:
\begin{eqnarray}
k>t:\nonumber\\
\!\!\!\!\!\!\!\!\!A^{N-k}_{i_{new}} \!\!\!\!\!& = &\!\!\!\!\! A^0_0
+ (D^{N-t-1}_{i_1\%2^{N-t-1}} +
\sum_{m=i_1+1}^{i_2-1}D^{N-t-1}_{m\%2^{N-t-1}} +
D^{N-t-1}_{i_2\%2^{N-t-1}}\nonumber\\
&& - d^{N-t-1}_{i_1\%2^{N-t-1}} +
d^{N-t-1}_{i_2\%2^{N-t-1}})/2^{k-t}
\nonumber\\
k \leq t: \nonumber\\
d^{N-k}_{i_{new}} & = & d^{N-k}_{(i+s/2^k)\%2^{N-k}}\nonumber\\
\mbox{where,}\nonumber\\
i_1&=&2^{k-t-1}i+\lfloor{s/2^{t+1}}\rfloor\nonumber\\
i_2&=&2^{k-t-1}(i+1)+\lfloor{s/2^{t+1}}\rfloor\label{eq:Afinal}
\end{eqnarray}

The above relation can now be used to evaluate the new blur
coefficients of a partially transformed signal with $k$ reduction
steps after any shift $s=0,...,2^N-1$ using only the coefficients of
the 0-shift signal. The worst case complexity for evaluating
$A^{N-k}_{i_{new}}$ using (\ref{eq:Afinal}) is \emph{O}$(\log(L))$,
where $L$ is the size of the signal $x(n)$ (see the complexity
analysis section for more details).

\section{Non-Integer Shifting}
In this section, we show how our solution can be extended to achieve
non-integer shifts. Although, our model is based on up-sampling the
original signal, the final relations that are derived require using
only the coefficients of the original signal. Up-sampling by a factor
of 2 can be modeled as adding levels to the lowest part of the
transform tree and setting the detail coefficients in those levels to
zero, with the lowest level being $N-1$. On the other hand, shifting
the up-sampled signal by an amount $u$ is equivalent to shifting the
original signal by $\frac{u}{2}$, which is a precision of
$\frac{1}{2}$. More generally, adding $h$ levels would enable
us to obtain a precision of $\frac{1}{2^h}$.

Let the size of the signal be $2^N$, $N'=N+h$ and $k=1+h,...,N+h$,
where $h$ is the number of added levels. Equation (\ref{eq:dfinal})
can now be modified to allow for non-integer shifting by a precision
of $\frac{1}{2^h}$ as follows:
\begin{eqnarray}
k>t:\nonumber\\
d^{N'-k}_{i_{new}} \!\!\!\!\!& = &\!\!\!\!\!
(D^{N'-t-1}_{i_1\%2^{N'-t-1}} +
2\sum_{i_1+1}^{i_2-1}{D^{N'-t-1}_{m\%2^{N'-t-1}}} \nonumber\\
& - &\!\!\!\! 2\sum_{i_2+1}^{i_3-1}{D^{N'-t-1}_{m\%2^{N'-t-1}}} -
D^{N'-t-1}_{i_3\%2^{N'-t-1}} \nonumber\\
& - & \!\!\!\! d^{N'-t-1}_{i_1\%2^{N'-t-1}} + 2d^{N'-t-1}_{i_2\%2^{N'-t-1}} - d^{N'-t-1}_{i_3\%2^{N'-t-1}})/2^{k-t}\nonumber\\
k \leq t: \nonumber\\
d^{N'-k}_{i_{new}} & = & d^{N'-k}_{(i+s/2^k)\%2^{N'-k}}\nonumber\\
\mbox{where,}\;\;\;\nonumber\\
i_1&=&2^{k-t-1}i+\lfloor{s/2^{t+1}}\rfloor\nonumber\\
i_2&=&2^{k-t-2}(2i+1)+\lfloor{s/2^{t+1}}\rfloor\nonumber\\
i_3&=&2^{k-t-1}(i+1)+\lfloor{s/2^{t+1}}\rfloor\nonumber\\
\mbox{and,   }\;\;\;\;&&\nonumber\\
d^{N'-t-1}_{i_2\%2^{N'-1}}&=&0 \mbox{, if $i_2$ is a
non-integer}\label{eq:dfinalNew2}
\end{eqnarray}

On the other hand, we can verify that $D^{N+h_0}_i =
D^N_{\lfloor{i/2^{h_0}}\rfloor}$, where $0\leq h_0\leq h$. Using
(\ref{eq:DDrel}), we also know that:
\begin{eqnarray}
D^N_i & = & D^{N-1}_{i/2}\;\;+\;d^{N-1}_{i/2}\;,\mbox{$i$ is even} \nonumber\\
& & {} D^{N-1}_{\lfloor{i/2}\rfloor}-d^{N-1}_{\lfloor{i/2}\rfloor},
\mbox{$i$ is odd} \label{eq:DDrelN}
\end{eqnarray}

The above result allows us to modify (\ref{eq:dfinalNew2}) in such a
way that avoids having to up-sample the signal for non-integer shifts,
saving thus memory space in actual implementation, especially that the
size increases exponentially. However, We have to split the equation
into two cases. The first is when $h\geq t+1$, which is when the
coefficients at the added levels are being used to evaluate
$d^{N'-k}_{i_{new}}$. The second is when $t$ is large enough for the
coefficients at the original levels of the tree to be used. This leads
to the new form of the phase shifting relation for non-integer values
as follows:
\begin{eqnarray}
h\geq t+1:& &\nonumber\\
d^{N'-k}_{i_{new}} \!\!\!\!\!& = &\!\!\!\!\!
(D^N_{\lfloor\frac{i_1\%2^{N'-t-1}}{2^{N'-t-t}}\rfloor} +
2\sum_{i_1+1}^{i_2-1}{D^N_{\lfloor\frac{m\%2^{N'-t-1}}{2^{N'-t-t}}\rfloor}} \nonumber\\
\!\!\!\!\!\!\!\!& - &\!\!\!\!\!\!\!\!
2\sum_{i_2+1}^{i_3-1}{\!\!\!D^N_{\lfloor\frac{m\%2^{N'-t-1}}{2^{N'-t-t}}\rfloor}}
\!\!\!\!-\!\!
D^N_{\lfloor\frac{i_3\%2^{N'-t-1}}{2^{N'-t-t}}\rfloor})/2^{k-t}\nonumber\\
h<t+1: & &\nonumber\\
k>t:\nonumber\\
d^{N'-k}_{i_{new}} \!\!\!\!\!& = &\!\!\!\!\!
(D^{N'-t-1}_{i_1\%2^{N'-t-1}} +
2\sum_{i_1+1}^{i_2-1}{D^{N'-t-1}_{m\%2^{N'-t-1}}} \nonumber\\
& - &\!\!\!\! 2\sum_{i_2+1}^{i_3-1}{D^{N'-t-1}_{m\%2^{N'-t-1}}} -
D^{N'-t-1}_{i_3\%2^{N'-t-1}} \nonumber\\
& - & \!\!\!\! d^{N'-t-1}_{i_1\%2^{N'-t-1}} + 2d^{N'-t-1}_{i_2\%2^{N'-t-1}} - d^{N'-t-1}_{i_3\%2^{N'-t-1}})/2^{k-t}\nonumber\\
k \leq t: & & \nonumber\\
d^{N'-k}_{i_{new}} & = & d^{N'-k}_{(i+s/2^k)\%2^{N'-k}}\nonumber\\
\mbox{where,}\;\;\;\nonumber\\
i_1&=&2^{k-t-1}i+\lfloor{s/2^{t+1}}\rfloor\nonumber\\
i_2&=&2^{k-t-2}(2i+1)+\lfloor{s/2^{t+1}}\rfloor\nonumber\\
i_3&=&2^{k-t-1}(i+1)+\lfloor{s/2^{t+1}}\rfloor\nonumber\\
\mbox{and,   }\;\;\;\;&&\nonumber\\
d^{N'-t-1}_{i_2\%2^{N'-1}}\!\!\!\!& = &0 \mbox{, if $i_2$ is a
non-integer}\label{eq:dfinalNew3}
\end{eqnarray}

The worst case complexity of the above formula is
\emph{O}$(\log(L+2^h))$ (again please refer to the Complexity Analysis
section for more details).

\section{N-Dimensional Shift}
Due to separability, an N-dimensional standard Haar transform is
constructed by applying the one-dimensional transform along each
dimension. As a result, the above solution can also be easily
generalized to N-dimensional signals by applying it along each
dimension separately.

\section{Complexity Analysis}
In this section we explain in further detail the complexity of
evaluating $d^{N-k}_{i_{new}}$ using equation (\ref{eq:dfinal}),
$A^{N-k}_{i_{new}}$ using equation (\ref{eq:Afinal}) and
$d^{N-k}_{i_{new}}$ using equation (\ref{eq:dfinalNew3}).

By examining (\ref{eq:dfinal}), it is easy to verify that the
complexity of evaluating $d^{N-k}_{i_{new}}$ can be expressed by the
difference of the bounds of the two sums in the equation, that is
\emph{O}($i_3-i_1$). Substituting the values for $i_1$ and $i_3$,
the complexity can be shown to be \emph{O}($2^{k-t-1}$) when $k>t$. When
$k\leq t$ the complexity becomes \emph{O}(1). Therefore, one can
determine that the worst case is when $t=0$, that is when the shift
is odd. In that case the complexity of computing $d^{N-k}_{i_{new}}$
becomes \emph{O}($2^{k-1}$). Let $L=2^N$ be the size of the signal,
then the number of the detail coefficients in a fully transformed
signal is $L-1=2^N-1$. At reduction level $k=N$, i.e. the root, the
complexity of evaluating $d^0_{0_{new}}$ is \emph{O}($2^{N-1}$) =
\emph{O}($\frac{L}{2}$) with a probability of $\frac{1}{L-1}$. At
the next reduction level $k=N-1$, the complexity is
\emph{O}($2^{(N-1)-1}$) = \emph{O}($\frac{L}{2^2}$) with a
probability of $\frac{2}{L-1}$. Table (\ref{tab:complxty}) shows the
complexity and its probability at each reduction level $k$.

\begin{table}[h]
\begin{center}
\begin{tabular}{|@{}c@{}|@{}c@{}|c|}
  \hline
    \multirow{2}{*}{\hspace*{1mm}Reduction Level\hspace*{1mm}} & \multirow{2}{*}{\hspace*{1mm}Complexity\hspace*{1mm}} &
    \multirow{2}{*}{Probability=$\frac{\mbox{Number of Coefficients at k}}{\mbox{Number of
    Coefficients}}$}\\
    & & \\ \hline\hline
    $k=N$   & $\emph{O}(\frac{L}{2})$   & $\frac{1}{L-1}$ \\\hline
    $k=N-1$ & $\emph{O}(\frac{L}{2^2})$ & $\frac{2}{L-1}$ \\\hline
    $k=N-2$ & $\emph{O}(\frac{L}{2^3})$ & $\frac{2^2}{L-1}$ \\\hline
    $k=N-3$ & $\emph{O}(\frac{L}{2^4})$ & $\frac{2^3}{L-1}$ \\\hline
    $:$     & $:$                       & $:$ \\\hline
    $k=1$   & $\emph{O}(\frac{L}{2^N})$   & $\frac{2^{N-1}}{L-1}$ \\\hline
  \end{tabular}
\end{center} \vspace*{1mm}
\caption{Table of the complexity and probability at each reduction
level $k$ for the one-dimensional detail coefficient
$d^{N-k}_{i_{new}}$.}\label{tab:complxty}
\end{table}\vspace*{-5mm}

By multiplying the complexities and the probabilities in table
(\ref{tab:complxty}) and summing them up, the average performance of
the worst case for evaluating $d^{N-k}_{i_{new}}$ is found to be
\emph{O}($\log(L)$).

By following a similar analysis and examining (\ref{eq:Afinal}), one
can find that the complexity for evaluating $A^{N-k}_{i_{new}}$ is
\emph{O}($\log(L)$) as well. Also, by examining (\ref{eq:dfinalNew3})
one can find that complexity for evaluating $d^{N-k}_{i_{new}}$
after a non-integer shift is \emph{O}($\log(L+2^h)$), where $h$ is
the number of levels added to achieve the shift.

\section{Experimental Validation}

We validate our results on the problem of accurate image rotation
using the decomposition of the rotation matrix described in
\cite{KiesewetterGraf,16579,167460,DBLP:journals/tip/UnserTY95}. The
choice of this application is driven by the fact that it allows us to
evaluate all aspects such as integer and non-integer shifts, and the
separability property.
\subsection{Image Rotation}

We implement rotation as a sequence of sheers using the following
factorization
\cite{KiesewetterGraf,16579,167460,DBLP:journals/tip/UnserTY95}:

\begin{eqnarray}
R(\theta) = \left[ \begin{array}{c c}
                            \cos(\theta)\;\; -\sin(\theta)\\
                            \sin(\theta)\;\; \;\;\;\cos(\theta) \\
                            \end{array}
 \right] = \left[ \begin{array}{c c}
                            1\;\; -\tan(\frac{\theta}{2})\\
                            0\hspace{1.2cm}1 \\
                            \end{array}
 \right] \nonumber\\\times \left[ \begin{array}{c c}
                            1\hspace{1.2cm} 0\\
                            \sin(\theta)\;\;\;\;\;\;1 \\
                            \end{array}
 \right] \times \left[ \begin{array}{c c}
                            1\;\; -\tan(\frac{\theta}{2})\\
                            0\hspace{1.2cm}1 \\
                            \end{array}
 \right]\label{eq:rotation}
\end{eqnarray}

\begin{figure}[h]
\begin{center}
\begin{tabular}{cc}
\includegraphics[width=35mm]{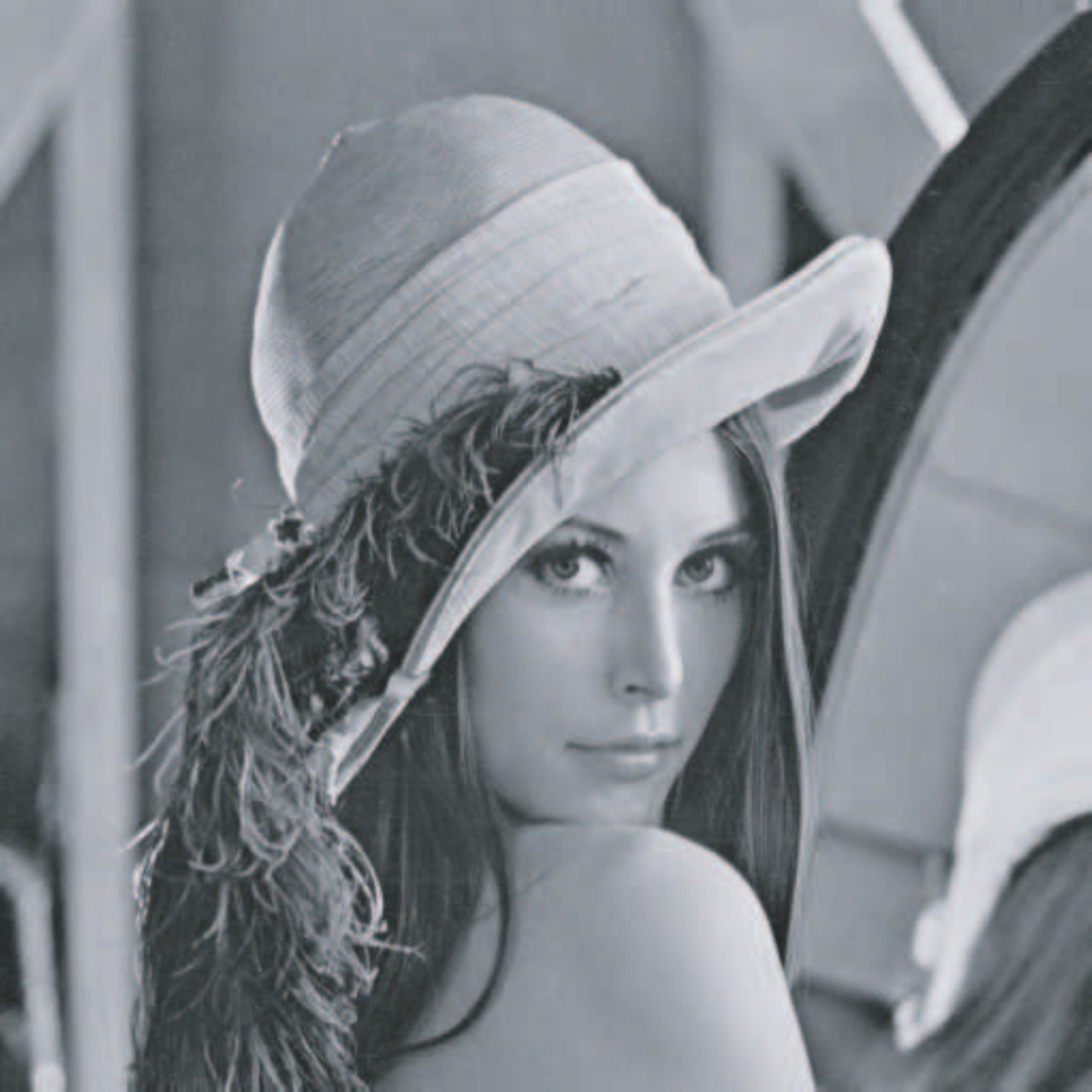} &
\includegraphics[width=48mm]{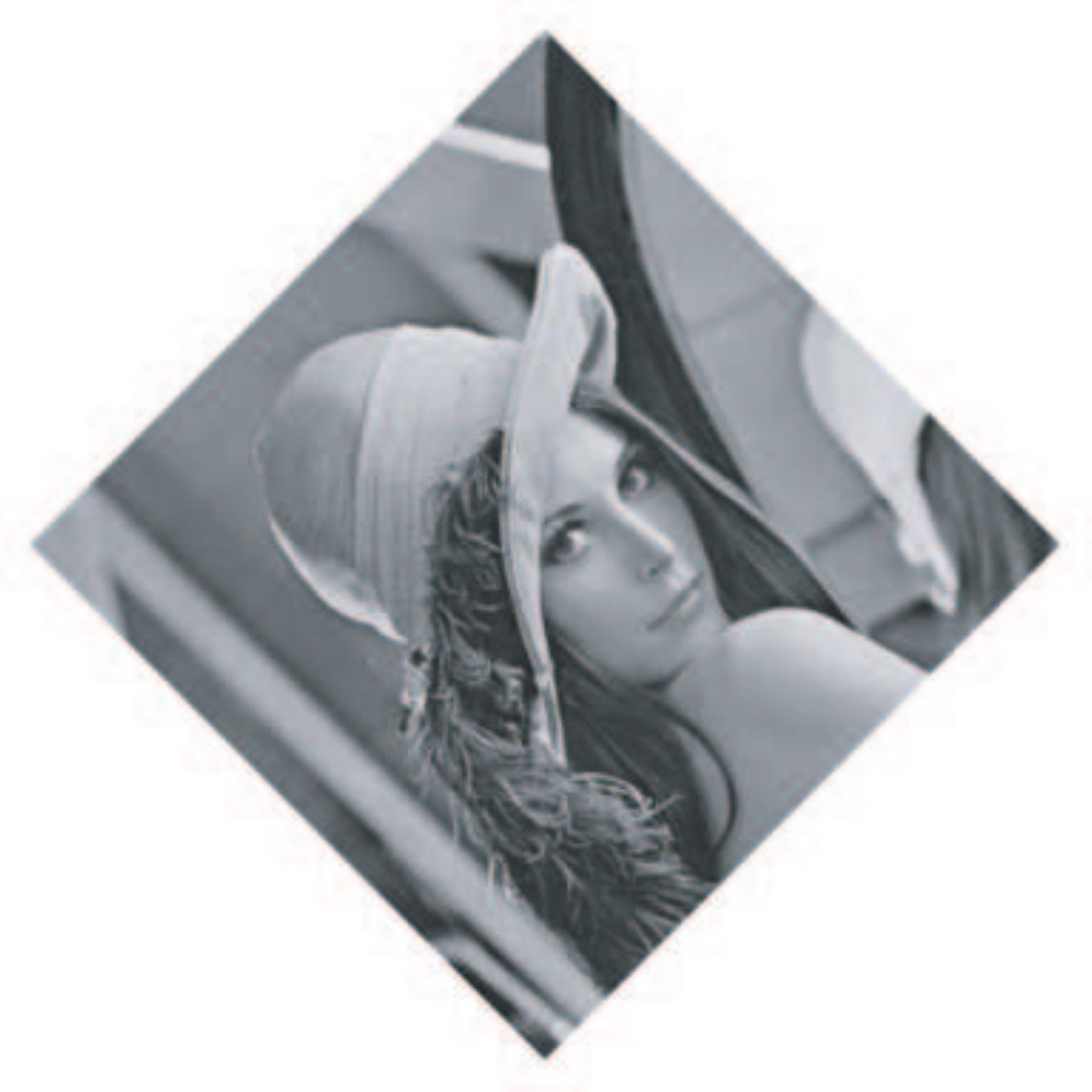}  \\
\end{tabular}
\end{center}
\caption{Original image, and the rotated one by $45$ degrees using
(\ref{eq:dfinalNew3}) with $h=3$.} \label{fig:rotation}
\end{figure}

\begin{figure}[h]
\begin{center}
\begin{tabular}{ccc}
\hspace*{-2mm}\includegraphics[width=28mm]{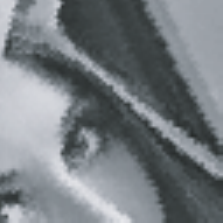}
&\hspace*{-3mm}
\includegraphics[width=28mm]{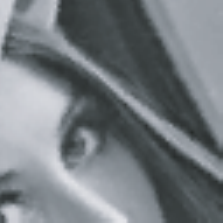} & \hspace*{-3mm}
\includegraphics[width=28mm]{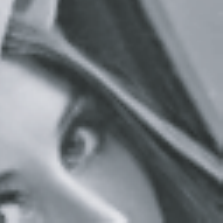} \hspace*{-3mm} \\
$h=0$ & $h=1$ & $h=2$  \\
\hspace*{-2mm}\includegraphics[width=28mm]{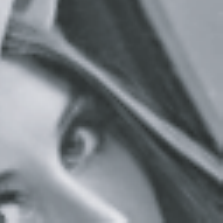}
&\hspace*{-3mm}
\includegraphics[width=28mm]{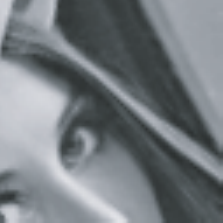} & \hspace*{-3mm}
\includegraphics[width=28mm]{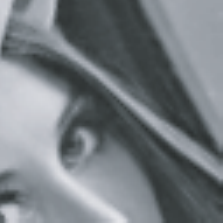} \hspace*{-3mm} \\
$h=3$ & $h=4$ & $h=5$  \\
\end{tabular}
\end{center}
\caption{A magnified portion of the image rotated using equation
(\ref{eq:dfinalNew3}). a. integer shift. b. non-integer shift with
precision of $\frac{1}{2}$. c. non-integer shift with precision of
$\frac{1}{2^2}$. d. non-integer shift with precision of
$\frac{1}{2^3}$. e. non-integer shift with precision of
$\frac{1}{2^4}$. f. non-integer shift with precision of
$\frac{1}{2^5}$.} \label{fig:rotationLevels}
\end{figure}

A sheer is in fact a sequence of shifts that are row-dependent, if the
sheer is horizontal, and column-dependent if it is vertical. That is,
each row is shifted by $\Delta x=-y\cdot\tan\frac{\theta}{2}$ in a
horizontal sheer while each column is shifted by $\Delta y=x
\cdot\sin\theta$ in a vertical sheer. Note that $\Delta x$ and $\Delta
y$ are in general non-integer values, hence, the applicability of our
phase-shifting relations derived in the previous sections. Figure
(\ref{fig:rotation}-b) shows the application of our method to the
3-step shearing image rotation with $h=3$. Figure
(\ref{fig:rotationLevels}) shows a magnified portion of the image
under different $h$ values.  An integer shift ($h=0$) results in a
jagged effect. This effect is eliminated, leading to higher quality
results, as we increase the value of $h$. Note that visually
satisfactory results are obtained even with $h=2$.

\begin{figure}[h]
\begin{center}
\begin{tabular}{ccc}
\hspace*{-2mm}\includegraphics[width=28mm]{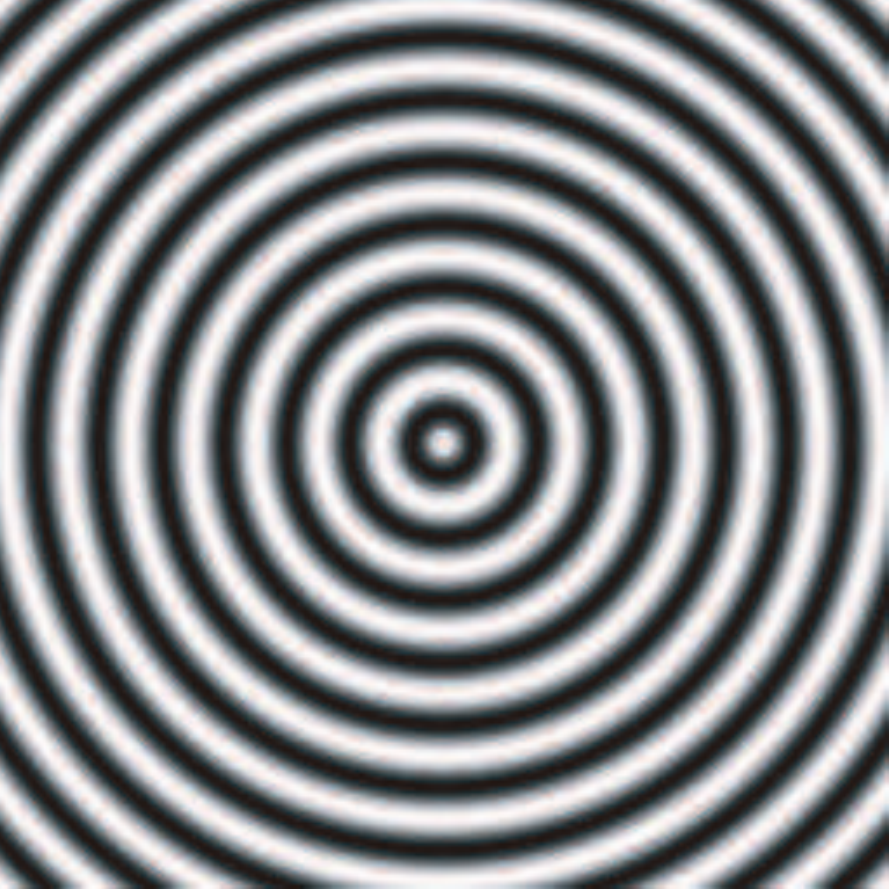}
&\hspace*{-4mm}
\includegraphics[width=28mm]{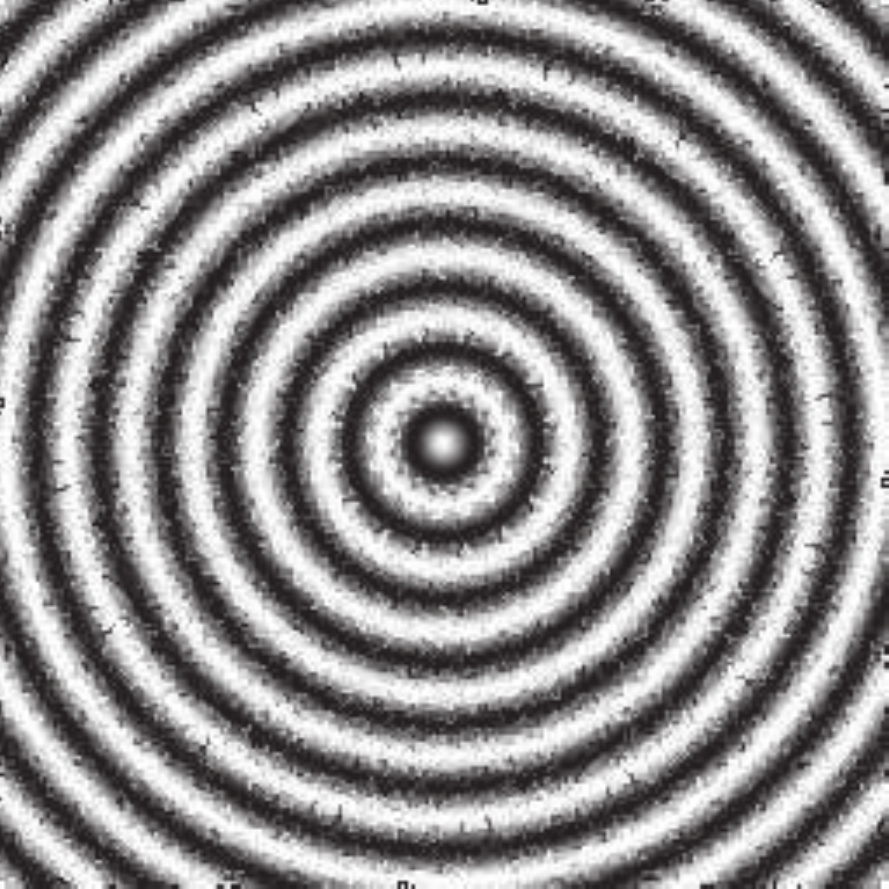} & \hspace*{-4mm}
\includegraphics[width=28mm]{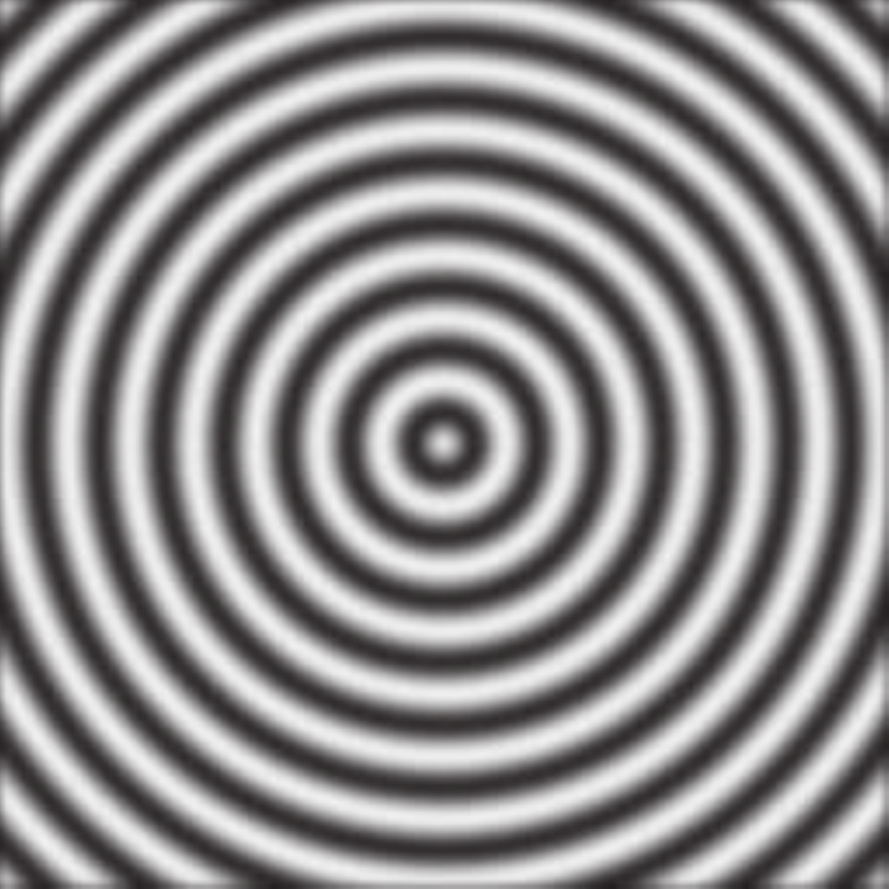} \hspace*{-4mm} \\
\small{original image} & \small{nearest neighbor}  & \small{bilinear} \\
\hspace*{-2mm}\includegraphics[width=28mm]{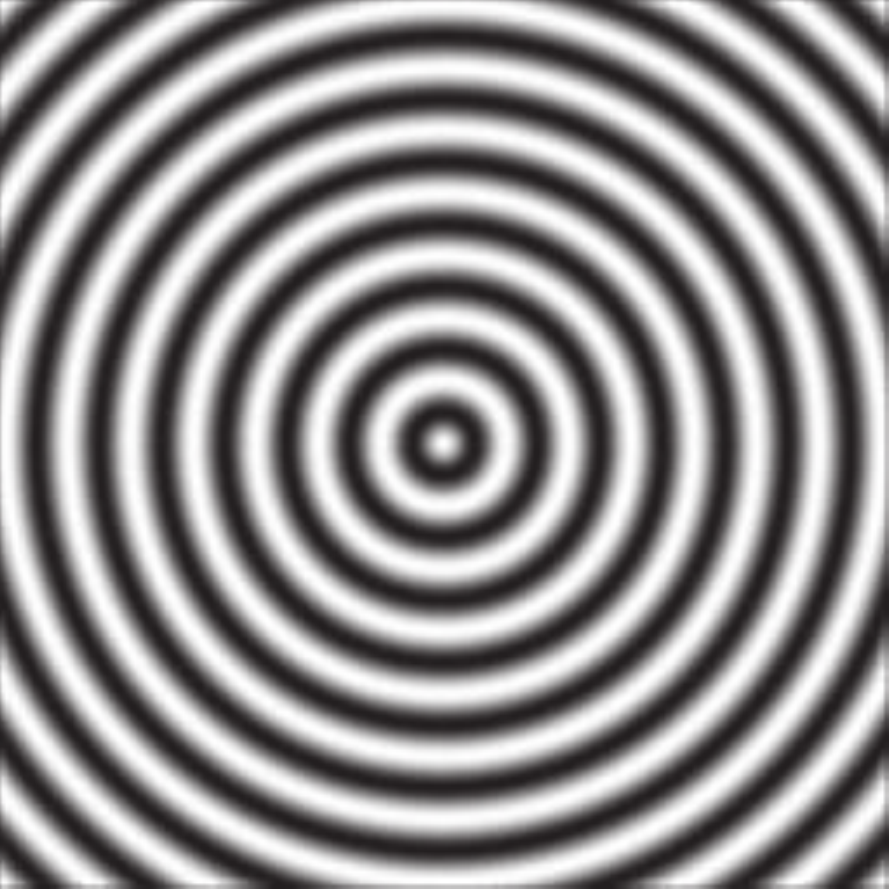}
&\hspace*{-4mm}
\includegraphics[width=28mm]{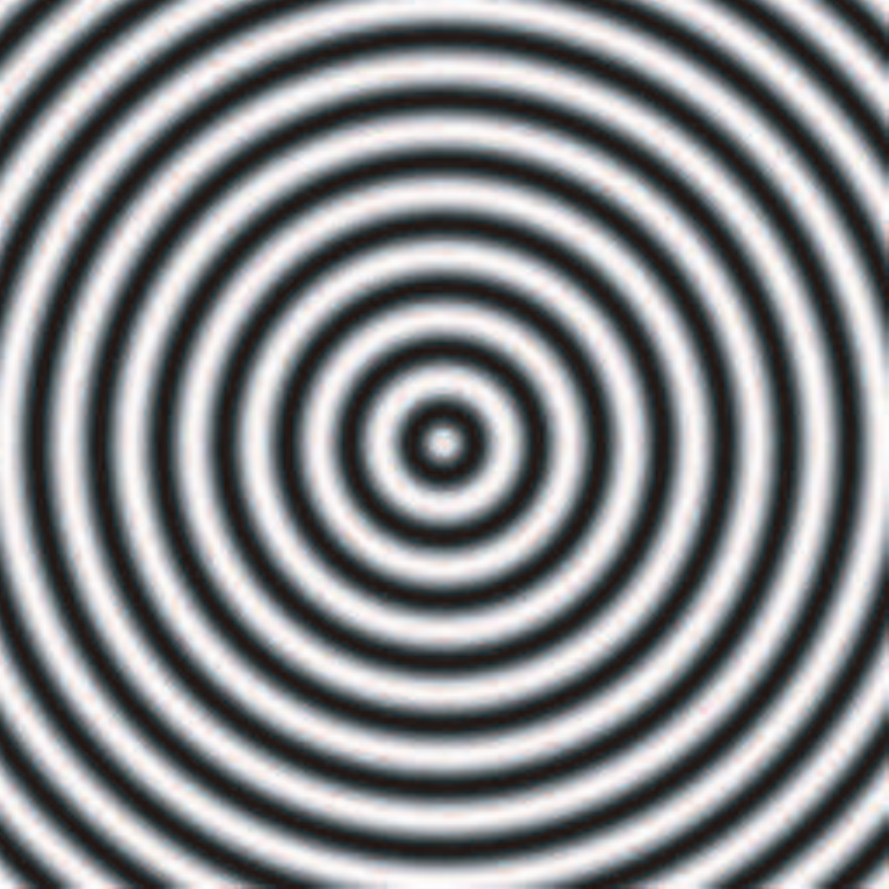} & \hspace*{-4mm}
\includegraphics[width=28mm]{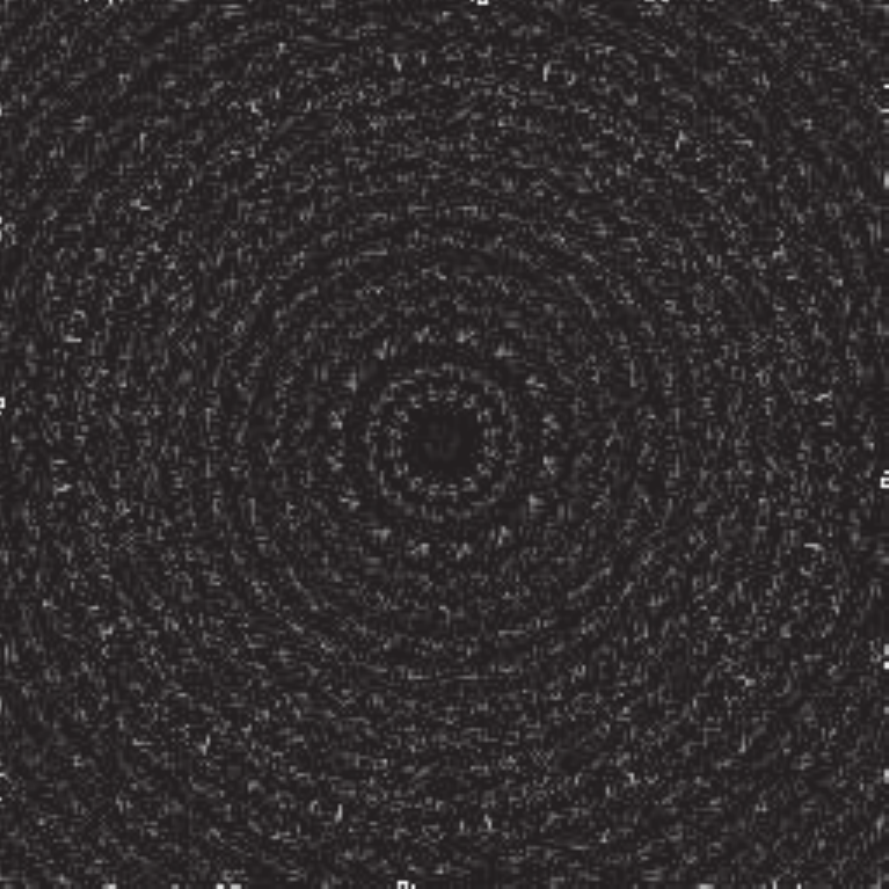} \hspace*{-4mm} \\
\small{bicubic} & \small{our method} & \small{nearest neighbor}  \\
&& \small{rms error = 28.95}\\
\hspace*{-2mm}\includegraphics[width=28mm]{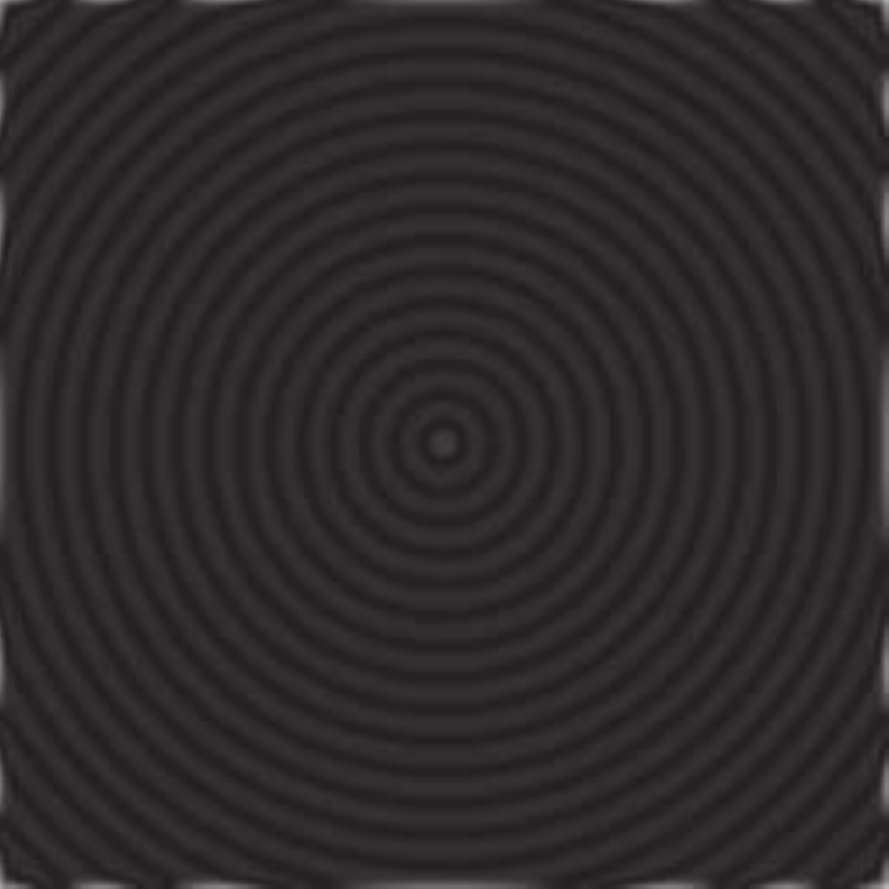}
&\hspace*{-4mm}
\includegraphics[width=28mm]{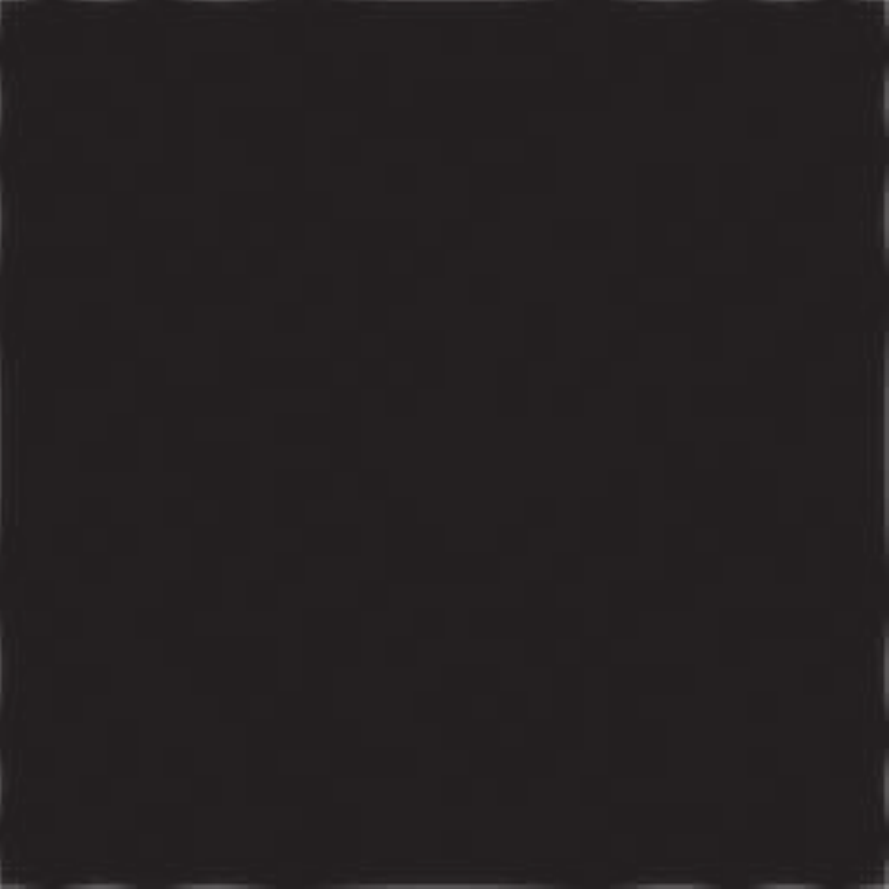} & \hspace*{-4mm}
\includegraphics[width=28mm]{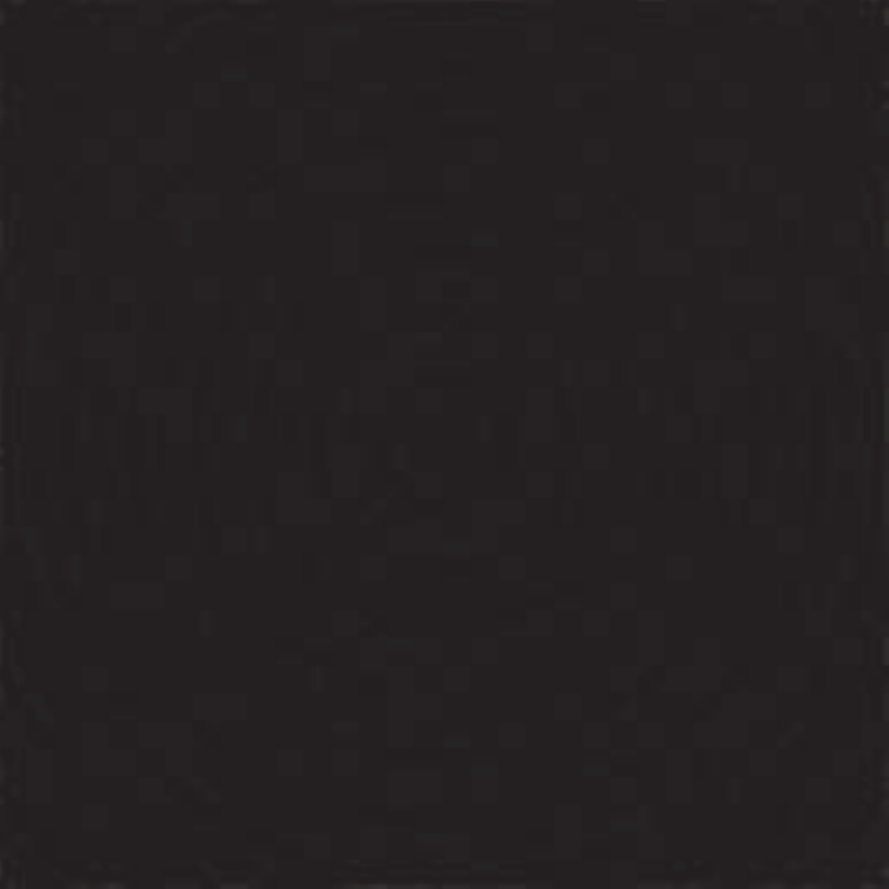} \hspace*{-4mm} \\
\small{bilinear} & \small{bicubic} & \small{our method}  \\
\small{rms error = 18.31} & \small{rms error = 7.38} & \small{rms error = 2.60}\\
\end{tabular}
\end{center}
\caption{The above images show the results of successively rotating
 the original image 16 times by a degree of $\frac{\pi}{8}$ for different methods including ours} \label{fig:comparisonC}
\end{figure}

\begin{figure}[h]
\begin{center}
\begin{tabular}{ccc}
\hspace*{-2mm}\includegraphics[width=28mm]{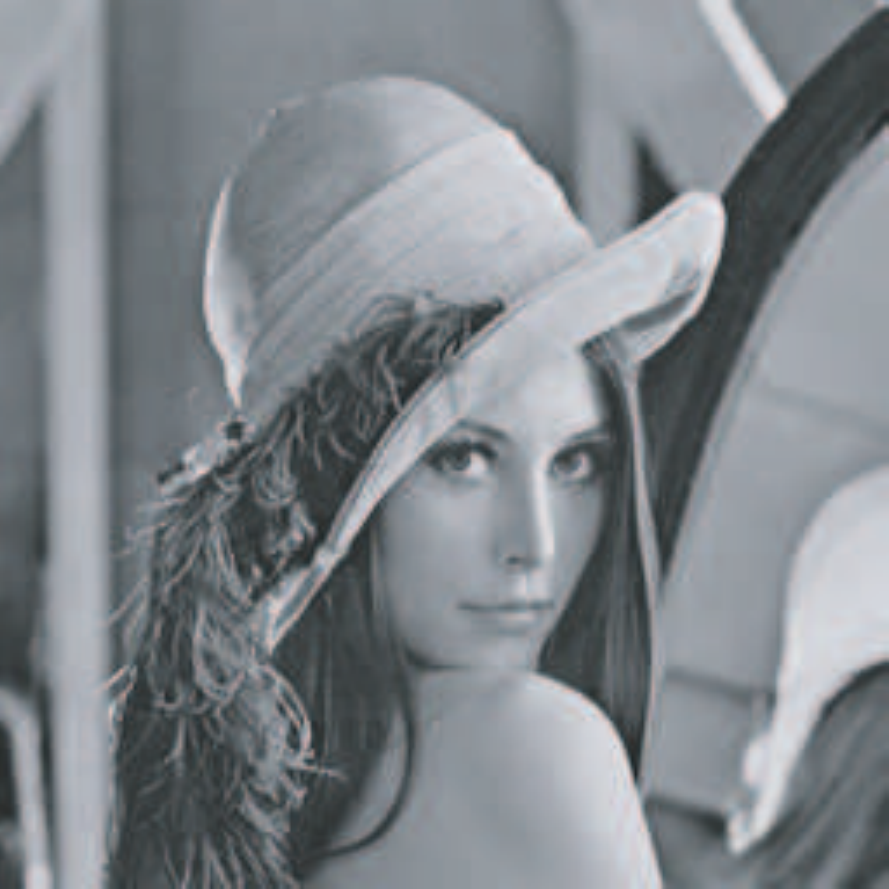}
&\hspace*{-4mm}
\includegraphics[width=28mm]{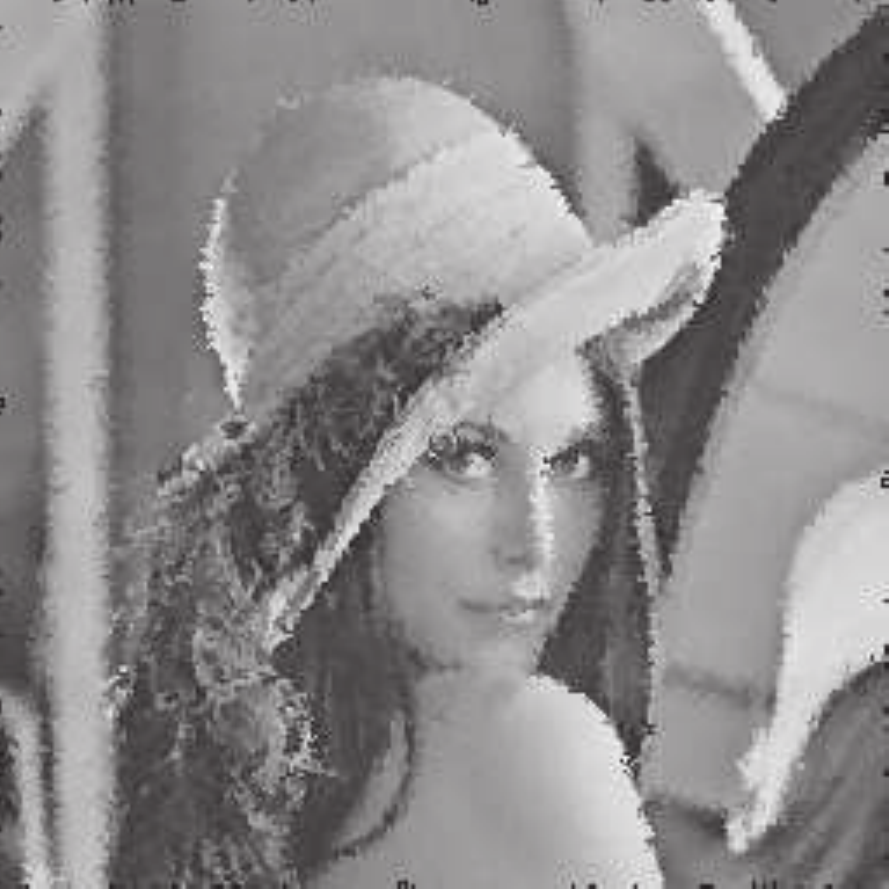} & \hspace*{-4mm}
\includegraphics[width=28mm]{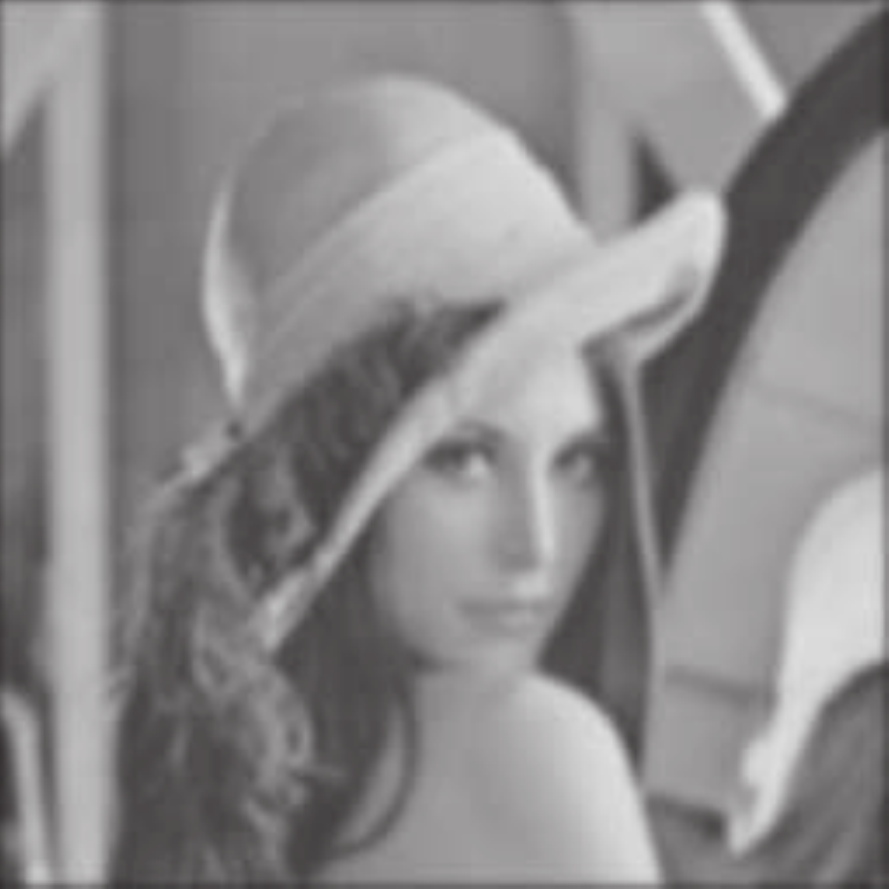} \hspace*{-4mm} \\
\small{original image} & \small{nearest neighbor}  & \small{bilinear} \\
\hspace*{-2mm}\includegraphics[width=28mm]{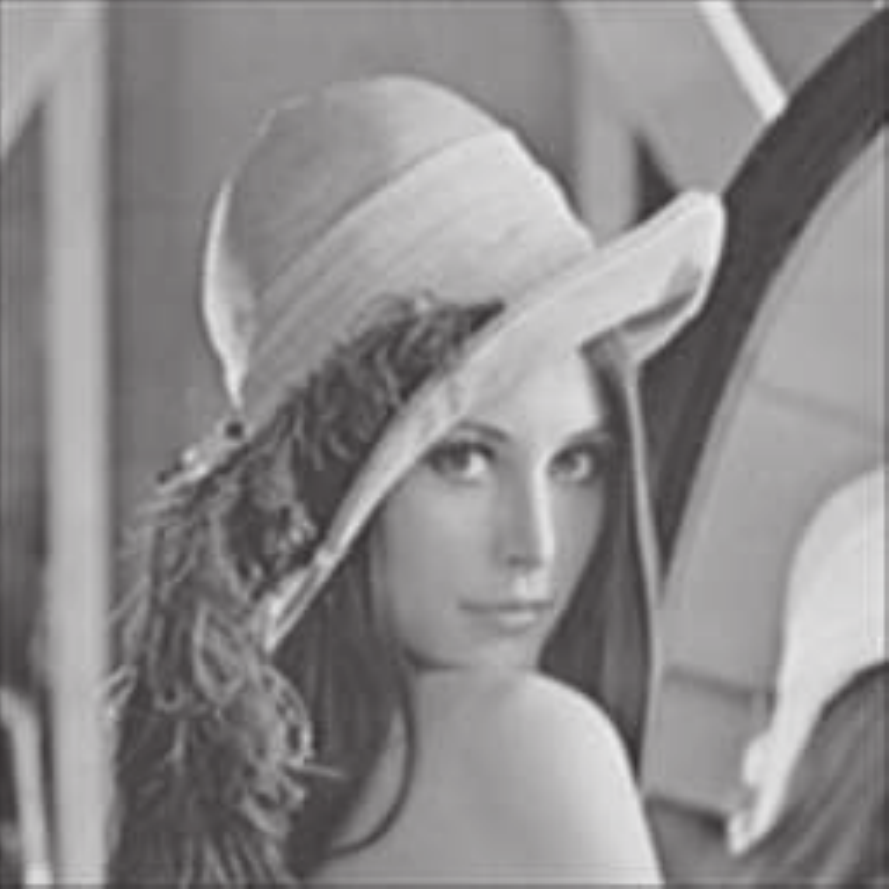}
&\hspace*{-4mm}
\includegraphics[width=28mm]{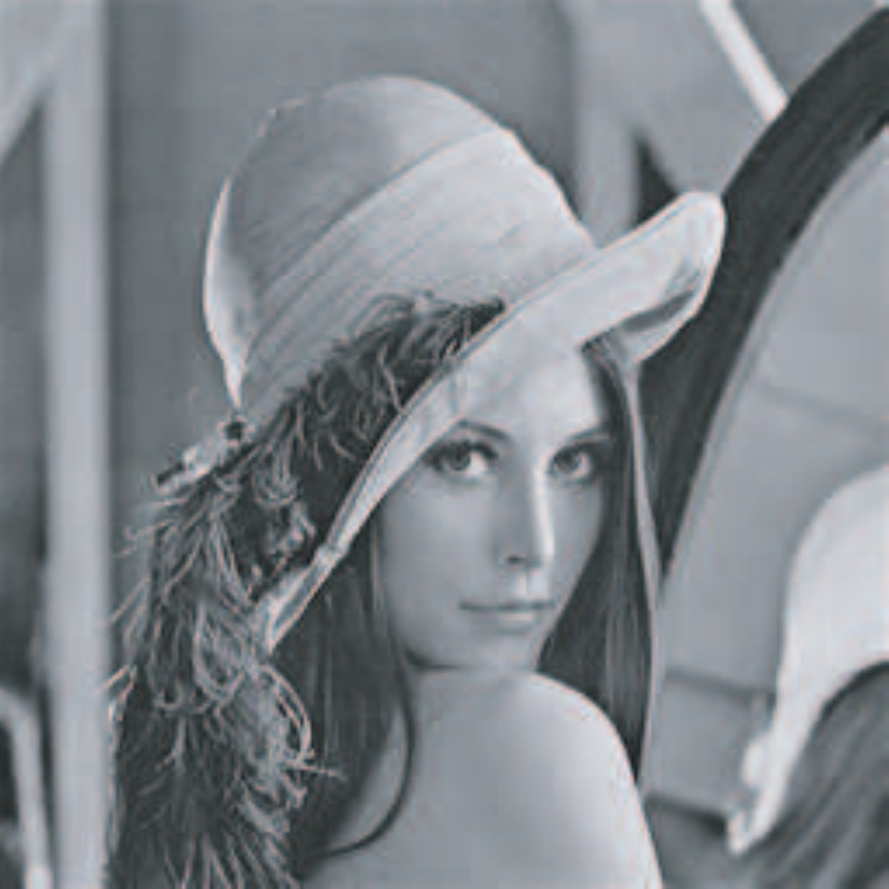} & \hspace*{-4mm}
\includegraphics[width=28mm]{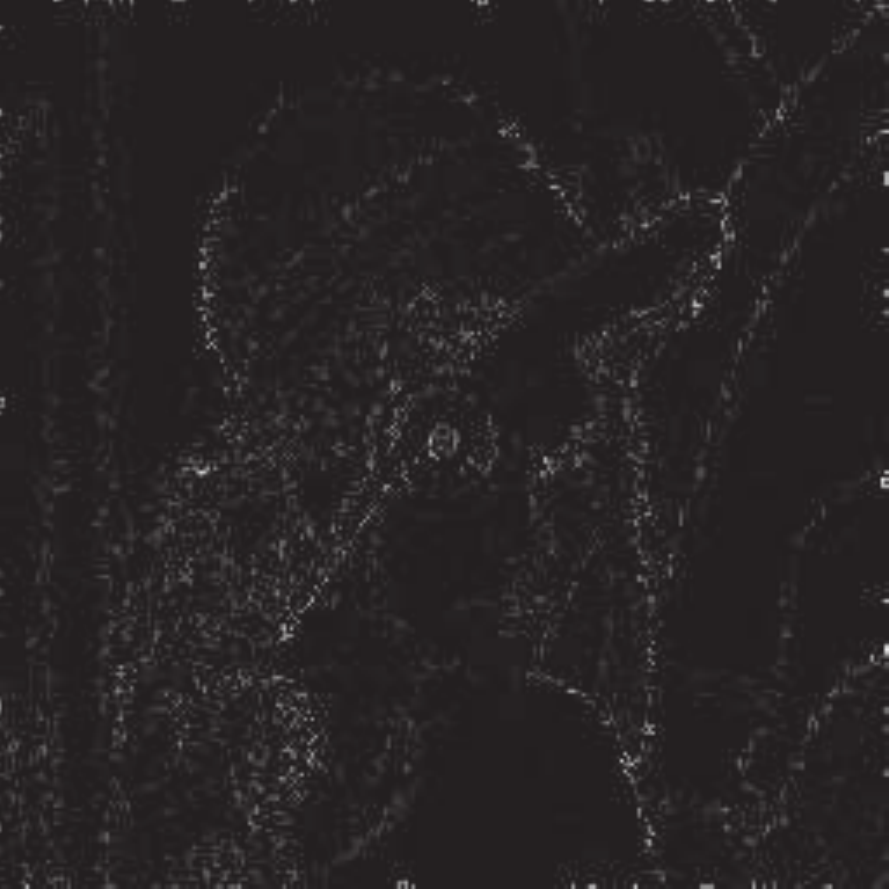} \hspace*{-4mm} \\
\small{bicubic} & \small{our method} & \small{nearest neighbor}  \\
&& \small{rms error = 15.93}\\
\hspace*{-2mm}\includegraphics[width=28mm]{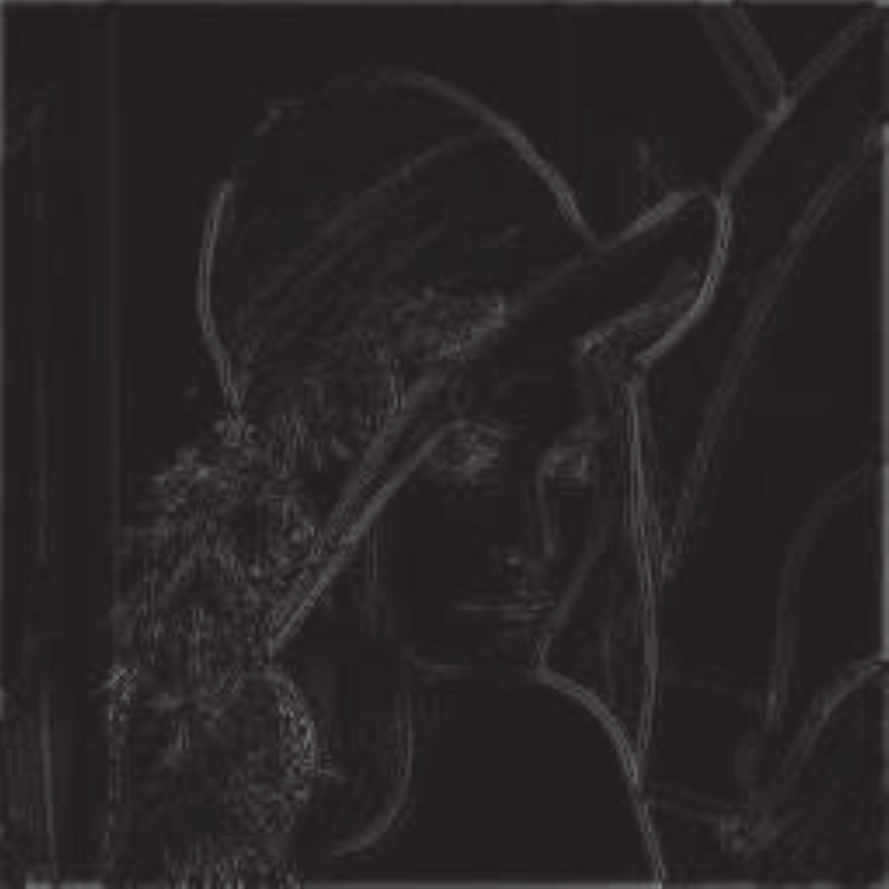}
&\hspace*{-4mm}
\includegraphics[width=28mm]{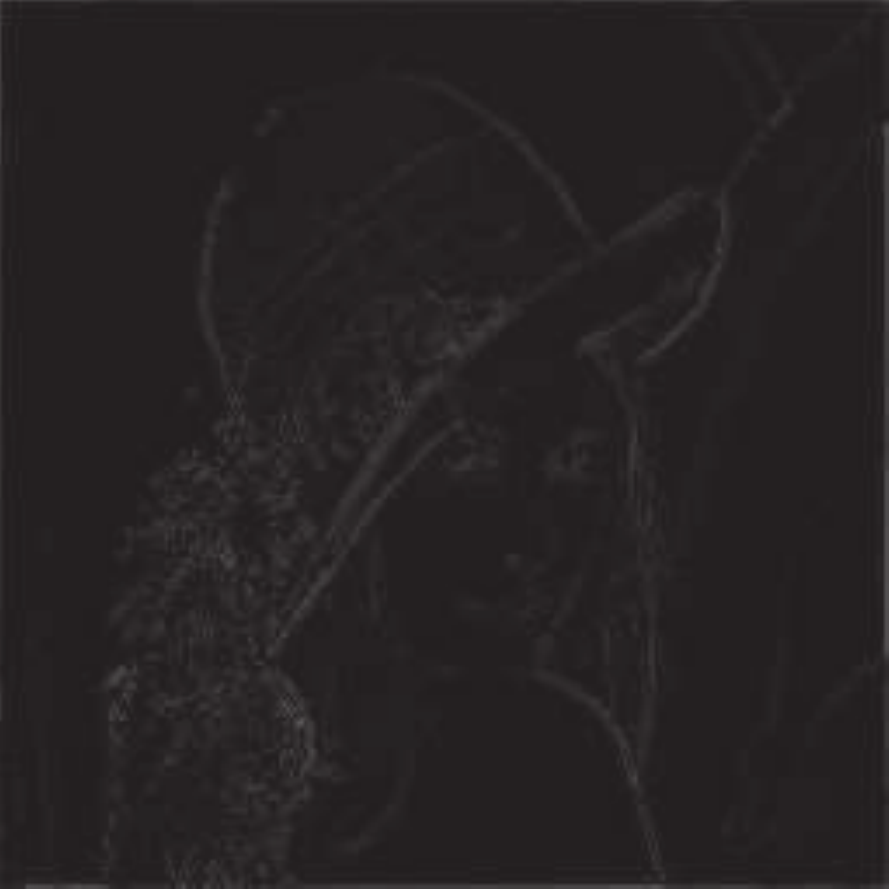} & \hspace*{-4mm}
\includegraphics[width=28mm]{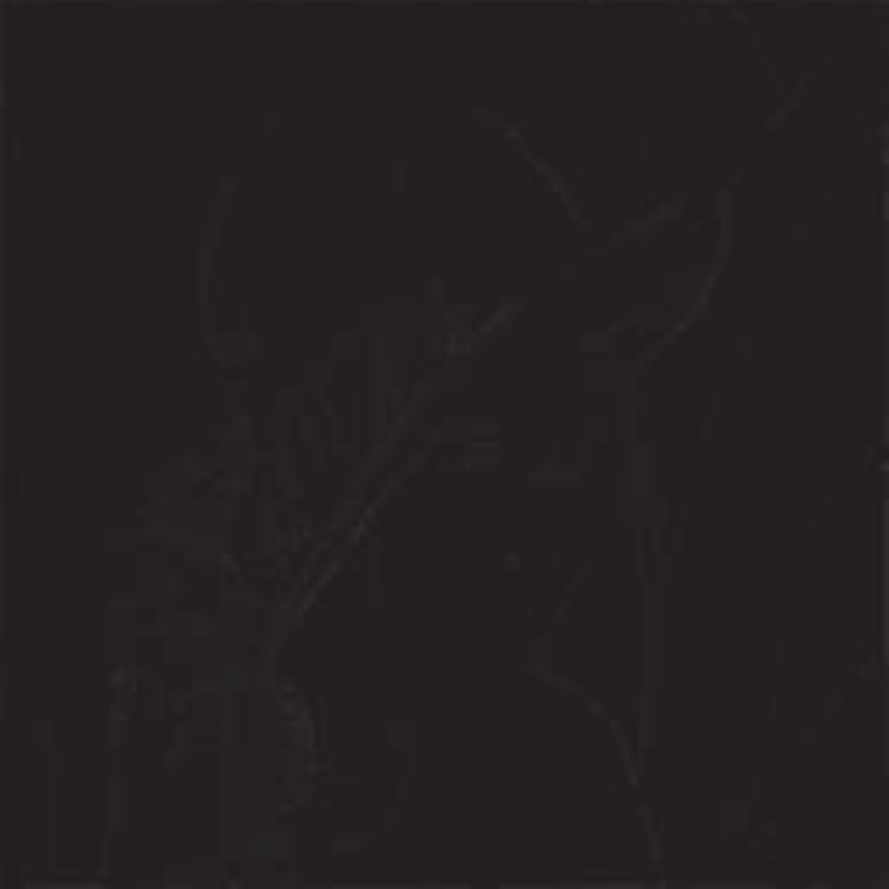} \hspace*{-4mm} \\
\small{bilinear} & \small{bicubic} & \small{our method}  \\
\small{rms error = 15.53} & \small{rms error = 9.08} & \small{rms error = 2.54}\\
\end{tabular}
\end{center}
\caption{The above images show the results of successively rotating
 the original image 16 times by a degree of $\frac{\pi}{8}$ for different methods including ours} \label{fig:comparisonL}
\end{figure}

As noted in \cite{DBLP:journals/tip/UnserTY95}, the worst scenario
occurs when the errors get accumulated. Therefore, in order to
quantify the performance, we computed the residual error, using an
experiment similar to the one adopted in
\cite{DBLP:journals/tip/UnserTY95}. In other words, we successively
rotated an input image by $\frac{\pi}{8}$ until it rotated back to its
original position. Figures \ref{fig:comparisonC} and
\ref{fig:comparisonL} show the results and the associated residual
errors on two standard test images for our method as compared to the
nearest-neighbor, bilinear, bicubic, and the sinc method. Note that
the image in Figure \ref{fig:comparisonC}, which was also used by
\cite{DBLP:journals/tip/UnserTY95}, is specificaly designed for
capturing accumulated errors in successive rotations. We tested and
compared our method extensively on many images, some of which are
shown in table \ref{tab:results}.


\section{Conclusion}

We have successfully shown that shift-invariance of the standard Haar
wavelets may be tackled directly by establishing analytic relations
between the Haar coefficients of a signal and its shifted
version. This new line of approach has the advantage that it does not
trade off the compression capability by retaining full decimation,
while preserving symmetry and separability. Our approach does not
yield a shift-invariant wavelet transform, but rather establishes the
explicit relations that describe phase-shifting directly in the
transform domain. Our experiments illustrate the validity of the
underlying motivating ideas, and the high accuracy of results in
practical problems.

\begin{table*}[t]
\begin{center}
\begin{tabular}{|c|c|c|c|c|c|c|c|c|c|c|}
\hline &\includegraphics[width=11mm]{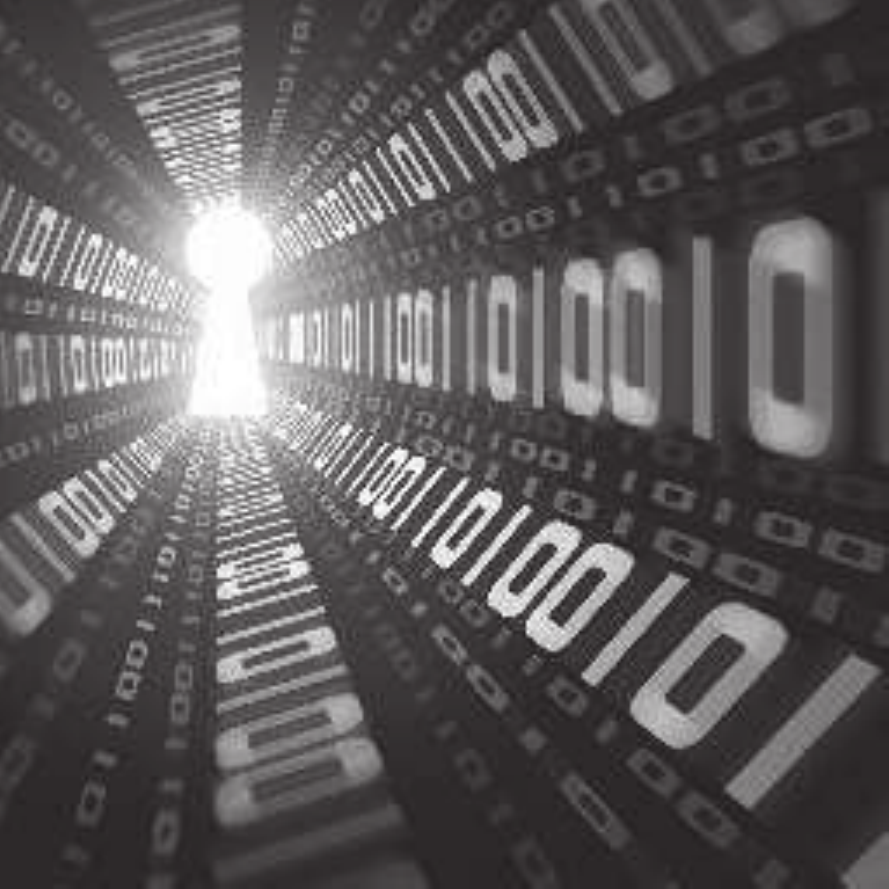}
&\includegraphics[width=11mm]{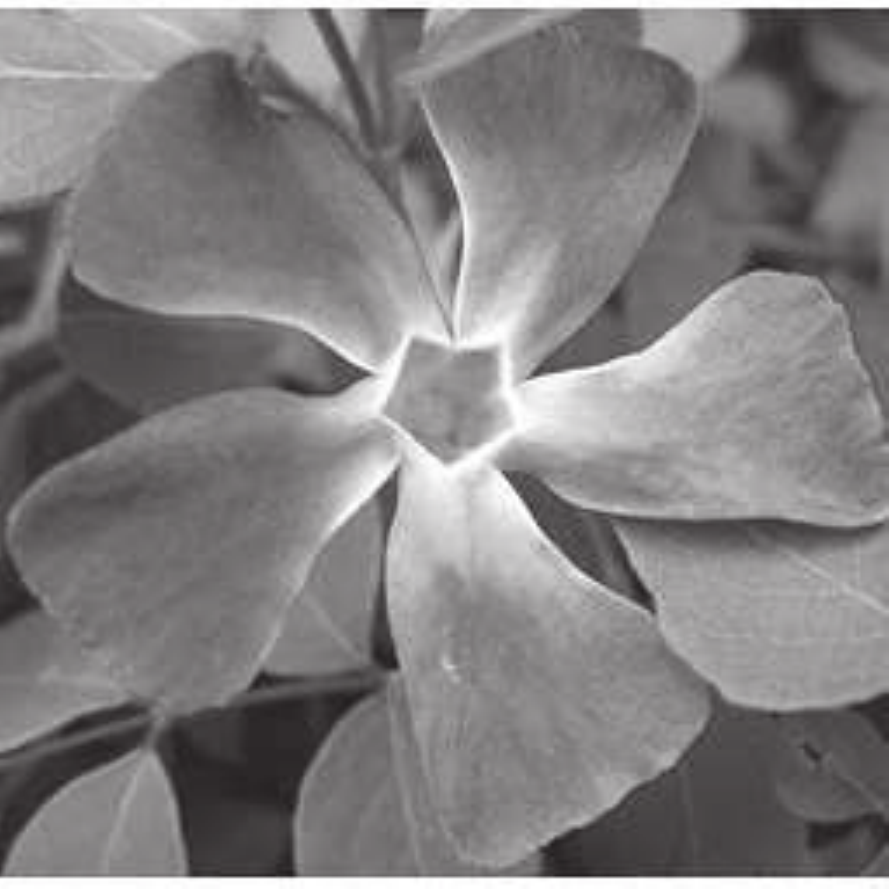}
&\includegraphics[width=11mm]{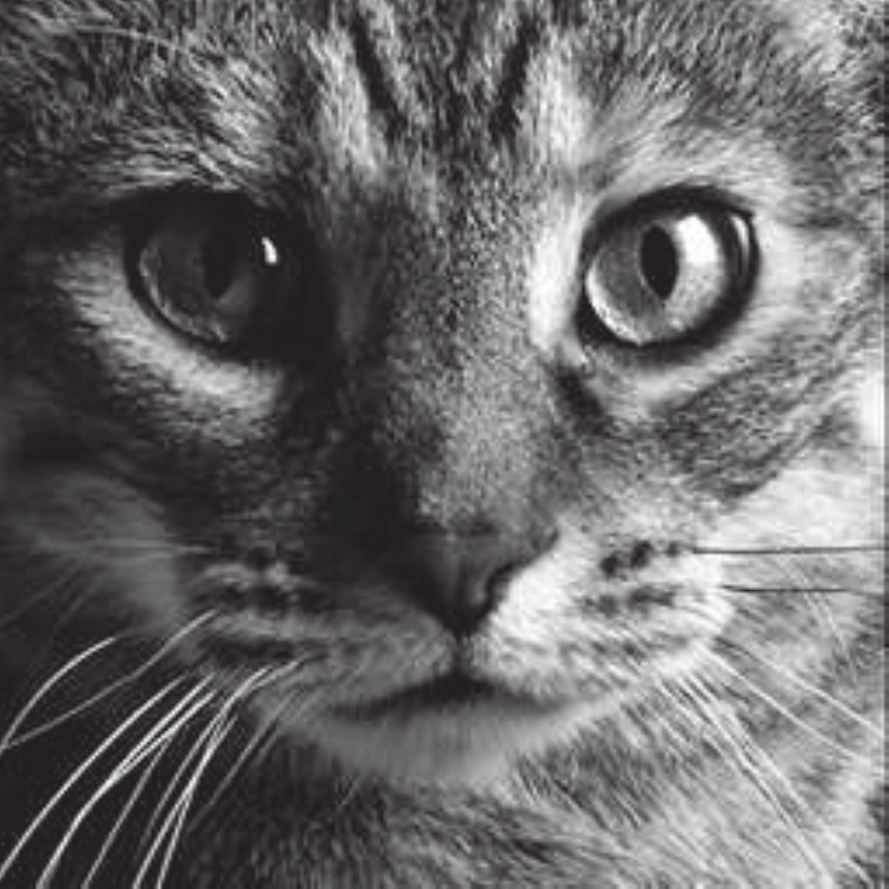}
&\includegraphics[width=11mm]{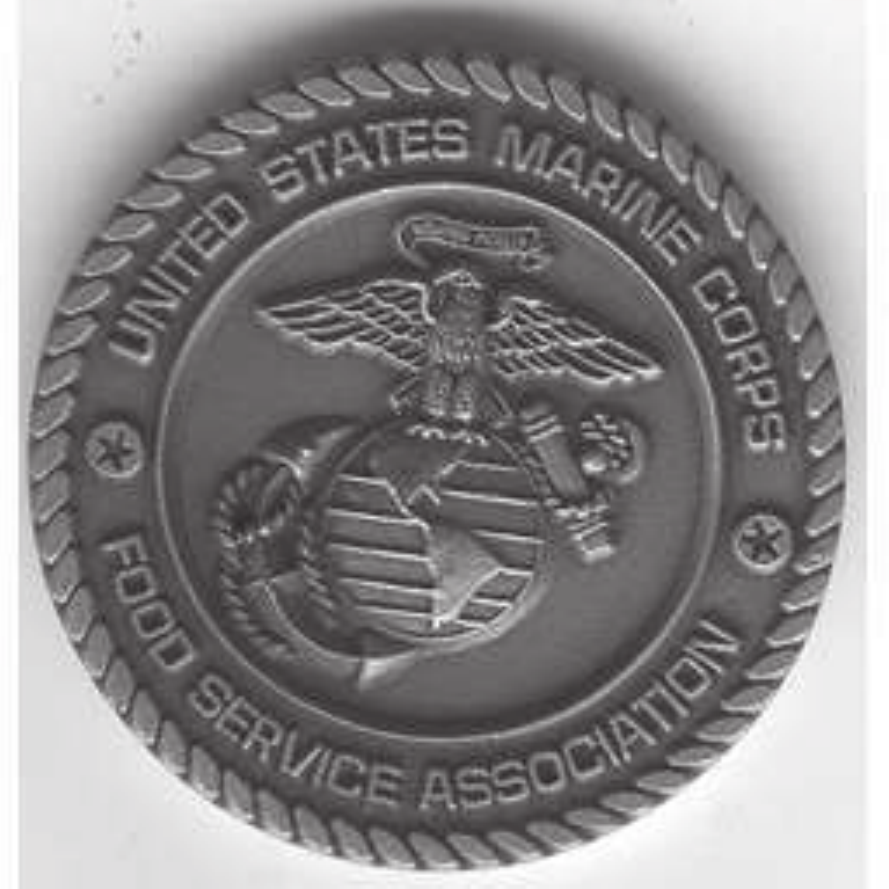}
&\includegraphics[width=11mm]{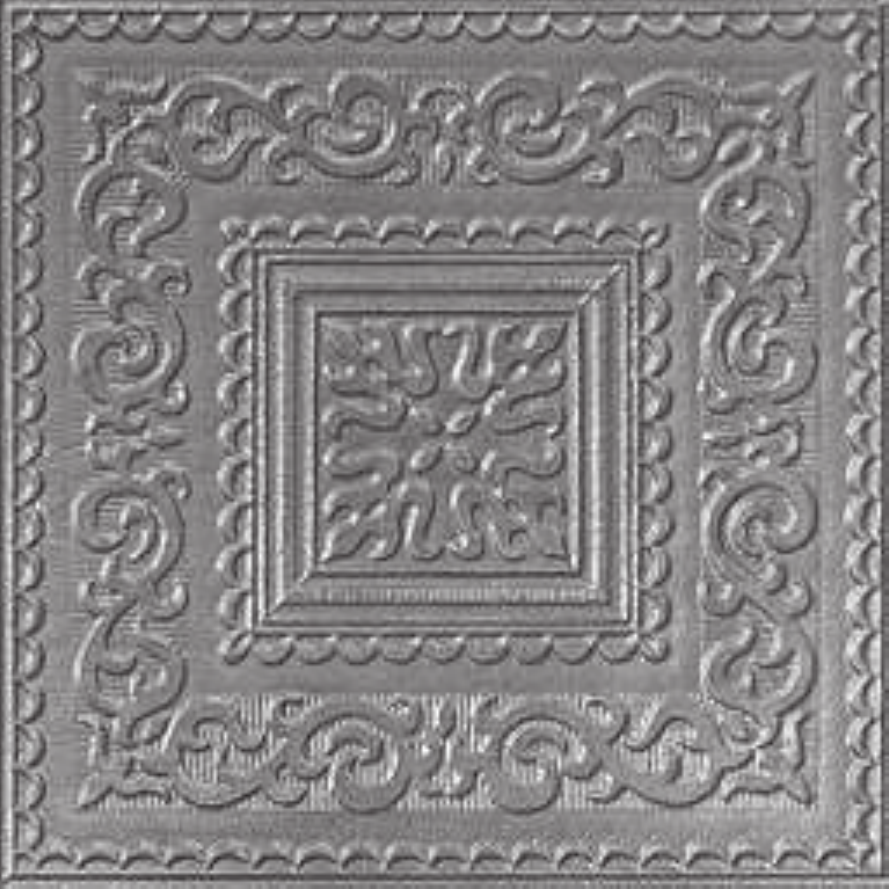}
&\includegraphics[width=11mm]{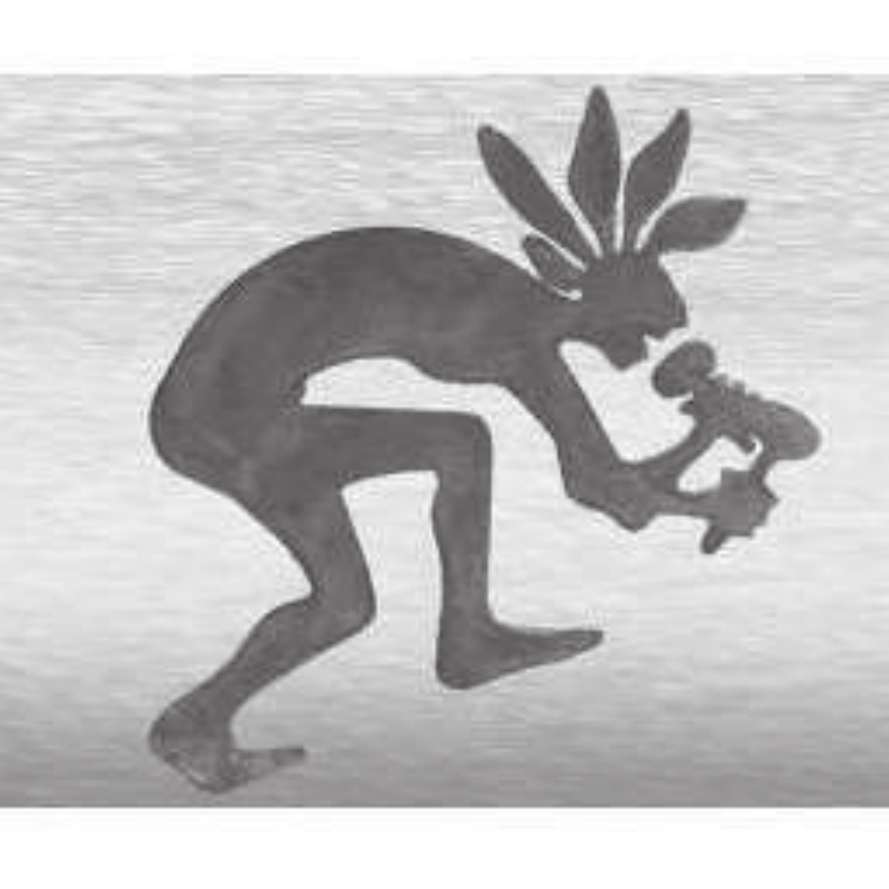}
&\includegraphics[width=11mm]{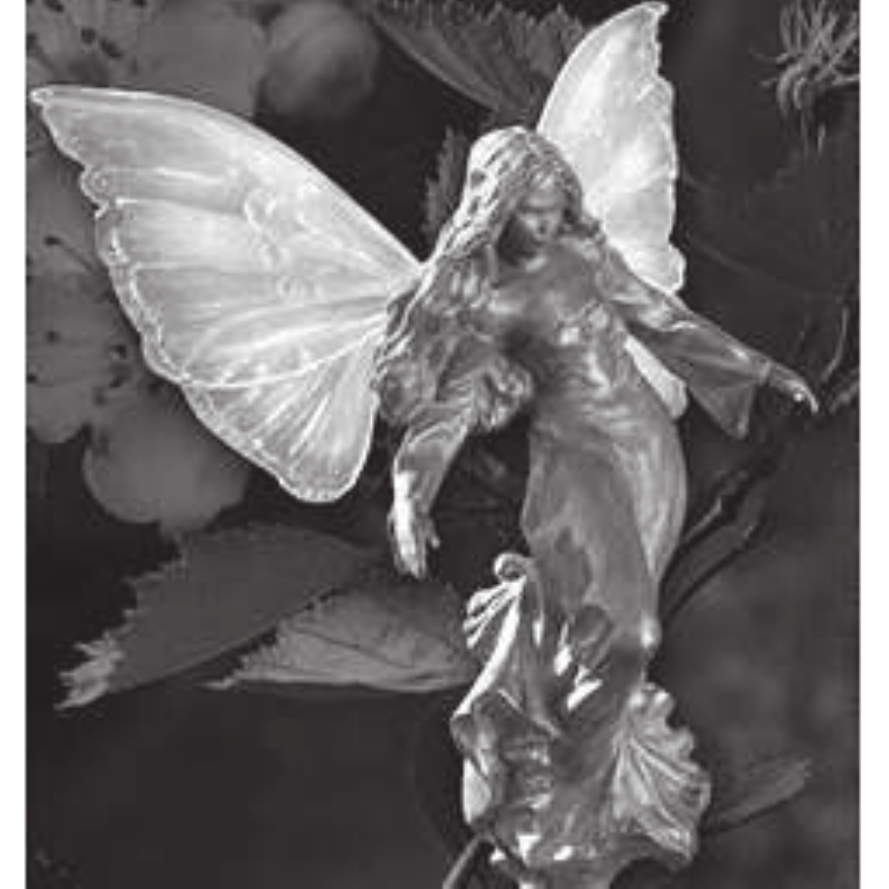}
&\includegraphics[width=11mm]{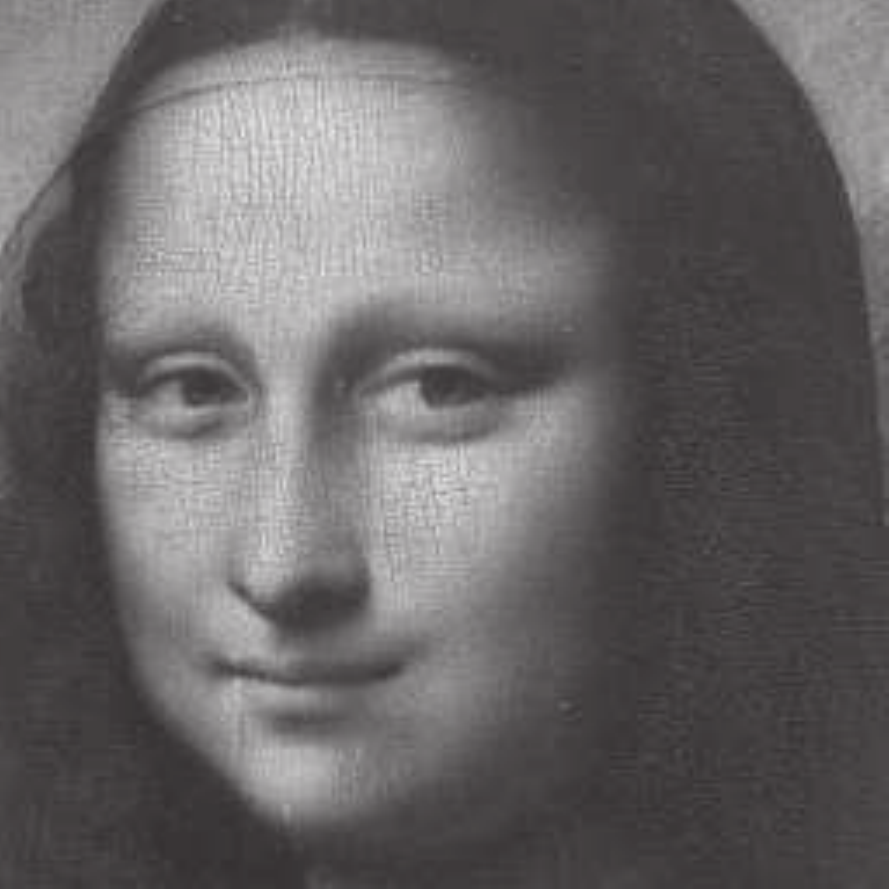}
&\includegraphics[width=11mm]{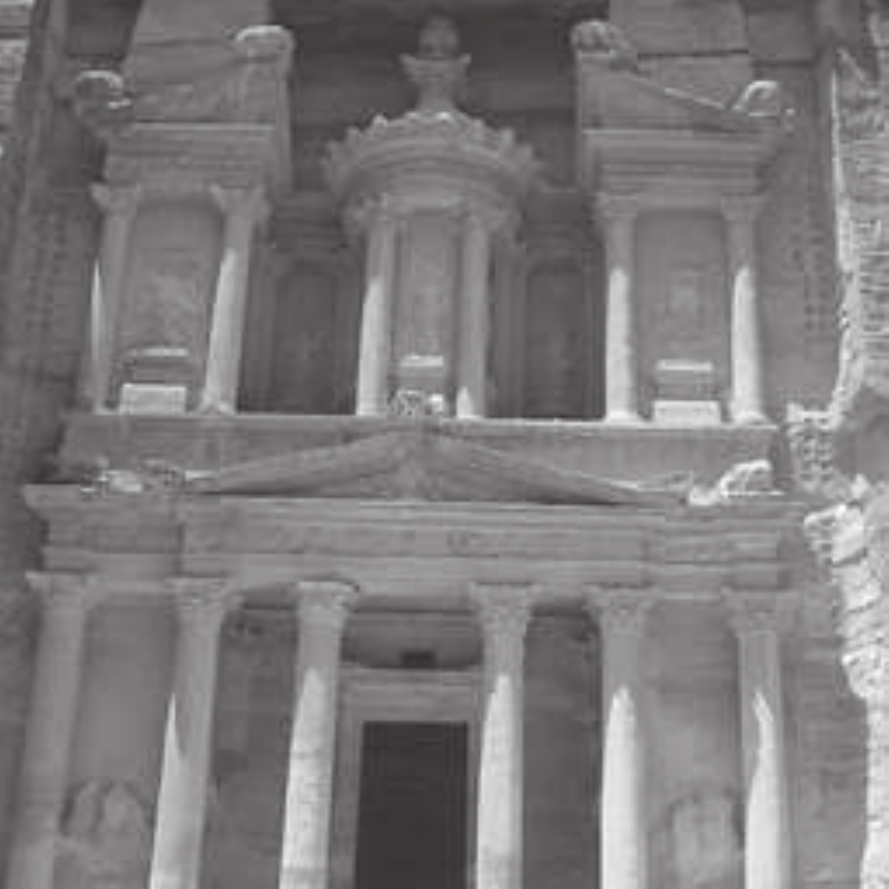}
&\includegraphics[width=11mm]{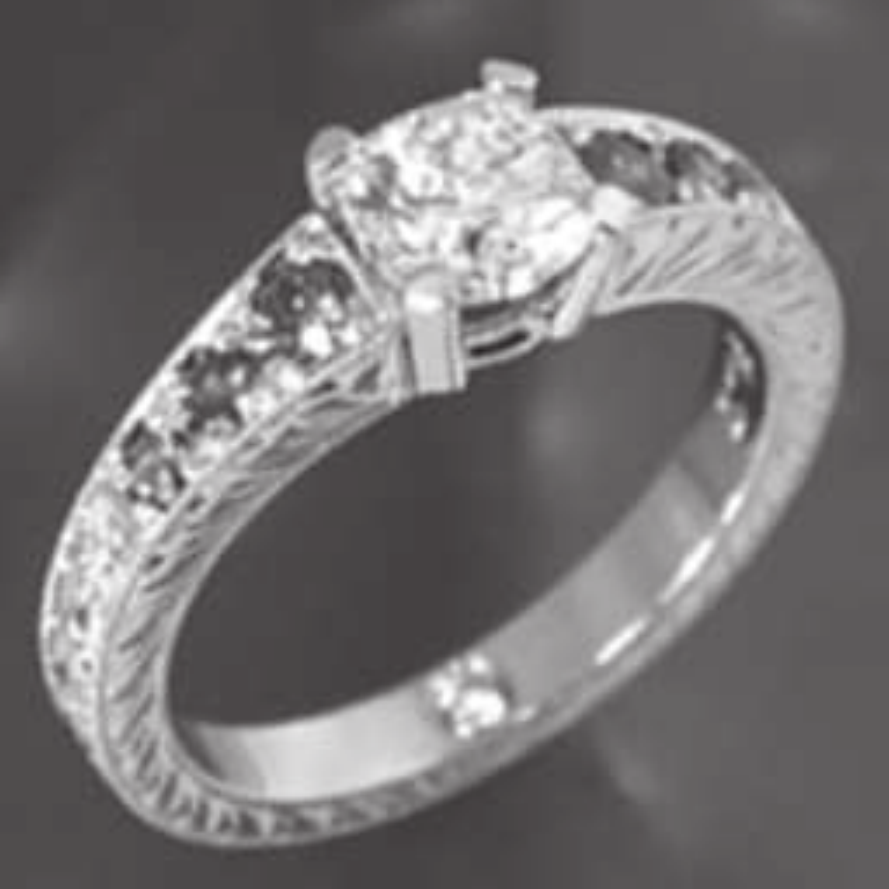}\\\hline \scriptsize{Nearest
Neighbor} & \scriptsize{23.2451} & \scriptsize{15.3117} & \scriptsize{23.2249} &
\scriptsize{19.3687} & \scriptsize{26.9128} & \scriptsize{13.8441} &
\scriptsize{22.1845} & \scriptsize{7.2702} & \scriptsize{11.8229} &
\scriptsize{18.3561}\\\hline \scriptsize{Bilinear} & \scriptsize{21.9343} &
\scriptsize{12.7582} & \scriptsize{21.9399} & \scriptsize{18.3487} &
\scriptsize{26.5292} & \scriptsize{12.0671} & \scriptsize{20.0879} & \scriptsize{6.2334}
& \scriptsize{10.3405} & \scriptsize{17.02}\\\hline \scriptsize{Bicubic} &
\scriptsize{15.2645} & \scriptsize{7.0842} & \scriptsize{14.5404} & \scriptsize{11.4509}
& \scriptsize{17.2482} & \scriptsize{6.7327} & \scriptsize{11.6671} &
\scriptsize{4.7183} & \scriptsize{5.9234} & \scriptsize{9.6489}
\\\hline \scriptsize{Sinc} &
\scriptsize{8.4349} & \scriptsize{1.8098} & \scriptsize{4.7284} & \scriptsize{4.0193} &
\scriptsize{6.4774} & \scriptsize{2.3743} & \scriptsize{2.9468} & \scriptsize{2.2373} &
\scriptsize{2.042} & \scriptsize{2.4533}
\\\hline
\scriptsize{Our Method} &  \scriptsize{3.3738} &  \scriptsize{1.5586} &
\scriptsize{3.0092} & \scriptsize{2.4095} & \scriptsize{3.4965} & \scriptsize{1.5173} &
\scriptsize{2.3753} & \scriptsize{1.1139} & \scriptsize{1.3243} & \scriptsize{2.0574}
\\\hline 
\end{tabular}
\end{center}
\vspace*{3mm}
\caption{Quantification and comparison of the accumulated residual error on several test images.} \label{tab:results}
\end{table*}

\bibliographystyle{plain}
\bibliography{foroosh,mais,barejournal}

\end{document}